\documentclass[twoside,11pt]{article}

\usepackage{jmlr2e}

\RequirePackage{amsmath}
\usepackage{algorithm}
\usepackage{algpseudocode}
\usepackage{indentfirst}
\usepackage{subcaption}

\usepackage{xcolor}

\newtheorem{thm}{Theorem}
\newtheorem{lem}{Lemma}
\newtheorem{cor}{Corollary}
\newcommand{\bx}{{\bf x}}

\newcommand{\wbox}{\sqcap\llap{$\sqcup$}}

\newcommand{\bmu}{\boldsymbol{\mu}}

\DeclareMathOperator*{\argmax}{arg\,max}

\ShortHeadings{Adjusted EI for Cumulative Regret in Noisy BO}{Hu, Wang, Dai, Low and Ng}
\firstpageno{1}

\begin{document}

\title{Adjusted Expected Improvement for Cumulative Regret Minimization in Noisy Bayesian Optimization}
%\title{Expected Improvement Adjustment for Cumulative Regret under Noisy Bayesian Optimization}

\author{\name Shouri Hu \email hushouri@u.nus.edu 
		 \AND
		\name Haowei Wang \email haowei\_wang@u.nus.edu 
		 \AND
		 \name Zhongxiang Dai \email dzx@nus.edu.sg
		 \AND
		\name Bryan Kian Hsiang Low \email lowkh@comp.nus.edu.sg
		\AND
		\name Szu Hui Ng \email isensh@nus.edu.sg
		\AND
       \addr 
       National University of Singapore}

\editor{ $\quad$ }

\maketitle

\begin{abstract}%   <- trailing '%' for backward compatibility of .sty file
%Bayesian optimization (BO) has been widely applied for global optimization of expensive black-box functions, and its performance depends highly on the sequential sampling strategy defined through an acquisition function. 
%Among the many acquisition functions proposed, the expected improvement (EI) is one of the most popular and has demonstrated good empirical performances in many applications for the minimization of simple regret. 
The expected improvement (EI) is one of the most popular acquisition functions for Bayesian optimization (BO) and has demonstrated good empirical performances in many applications for the minimization of simple regret. 
However, under the evaluation metric of cumulative regret, the performance of EI may not be competitive, and its existing theoretical regret upper bound still has room for improvement.
To adapt the EI for better performance under cumulative regret, we introduce a novel quantity called the \emph{evaluation cost} which is compared against the acquisition function,
and with this, develop the \emph{expected improvement-cost} (EIC) algorithm.
In each iteration of EIC,  a new point with the largest acquisition function value is sampled, only if that value exceeds its evaluation cost.
If none meets this criteria, the current best point is resampled.
This evaluation cost quantifies the potential downside of sampling a point, which is important under the cumulative regret metric as the objective function value in every iteration affects the performance measure. 
We establish in theory a high-probability regret upper bound of EIC based on the maximum information gain, which is tighter than the bound of existing EI-based algorithms.
It is also comparable to the regret bound of other popular BO algorithms such as Thompson sampling (GP-TS) and upper confidence bound (GP-UCB).
We further perform experiments to illustrate the improvement of EIC over several popular BO algorithms.

\end{abstract}


\section{Introduction}
\label{section:introduction}
%Introduce BO framework
Bayesian optimization (BO) is a sequential design framework for the global optimization of black-box functions with the following key characteristics:
Firstly, the (objective) function has an unknown structure, with the input vectors residing in low- to moderate-dimensional Euclidean space.
Secondly, the evaluation of such functions is expensive, hence it is impossible to search the entire domain in high precision before exhausting the budget.
Thirdly, the derivative information of the objective function is unavailable or impractical to estimate, 
therefore classical gradient-based methods are not applicable.
BO was initially studied by \cite{Kushner64}, \cite{Mockus75}, \cite{Zilinskas75} and \cite{MTZ78},
and later popularized by the work of \cite{JSW98}.
Recently it has gained substantial attention in a large number of important areas such as engineering systems optimization and hyper-parameter tuning for machine learning algorithms \citep{TSDB18,KMHIK19,LKOB19,SZLJ21}.

%Introduce GP model
Within the BO framework, the evaluated points are sequentially selected by maximizing an \emph{acquisition function}, whose calculation requires a surrogate function to model the objective function using the sequentially collected points and their (either noisy or noise-free) observations. 
The most commonly used surrogate functions are the Gaussian processes (GPs).
A GP is specified a priori by the mean function and the covariance kernel.
Given a set of training data, the posterior of a GP remains a GP, with closed-form expressions for the posterior distribution at any point.
This nice property makes the GPs become a powerful tool for statistical modelling.
Refer to \cite{WR06}, \cite{OGR09} and \cite{MSG16} for a more detailed introduction to GPs.

%Introduce first type of pratitioners
BO has been extensively applied in numerous fields, and the goal of many of them is to find the best final solution, i.e., to find the point that maximizes the objective function.
Under this goal, points with good chances to achieve the maximum function value are evaluated sequentially, and after the budget is exhausted, the evaluated point with the largest observed function value is usually reported as the final solution.
Application fields with this goal include engineering system optimization \citep{JSW98,TSDB18},
materials science design \citep{FW16,Packwood17,FHHM19},
and pharmaceutical product development \citep{BOA17,SKTK20}.
This goal in BO is typically known as the minimization of the \emph{simple regret}, and there has been several theoretical results on its properties in the literature \citep{GAOS10,Bull11,Ryzhov16,WSK21}.

%Introduce EI. Mention KG and ES in literature
To minimize the simple regret of BO, a number of acquisition functions have been proposed.
Among them, one of the most widely used acquisition functions is \emph{expected improvement} (EI).
EI was firstly proposed by \cite{Mockus75} under the noise-free BO setting, and then received further attention due to the work of \cite{JSW98} who successfully incorporated GPs into the calculation of EI.
As a conceptually intuitive method, EI has demonstrated impressive empirical performances in various applications.
In every iteration, EI calculates the expected gain (over the best observed function value so far) from every point in the domain based on the posterior mean and variance of the GP model, and evaluates the point that maximizes this expected gain.
Besides EI, other popular acquisition functions for simple regret minimization include \emph{knowledge gradient} which evaluates the point that maximizes the increment of posterior mean function \citep{FPD09,WF16}, and \emph{entropy search} which selects the point that is most informative about the location of the global optimum \citep{HS12,HHG14,WJ17}.

%%Motivate the idea of cumulative regret
Despite the popularity of BO methods based on simple regret minimization, many common applications of BO are also concerned with the overall performance of BO throughout the entire experiment, instead of only finding the best final solution (i.e., minimizing the simple regret).
One representative example of such applications is the recommendation system \citep{KBV09,KBKTC15,GGCPA20}, where algorithms (such as matrix-factorization) are deployed to recommend items (e.g., news, movies and songs) to customers
in order to increase their stickiness or likelihood to make a purchase.
Like many machine learning algorithms, these recommendation algorithms require a hyper-parameter tuning process in order to enhance their effectiveness, for which BO is usually a prominent choice.
The performance of recommendation algorithms, which is summarized as a numerical score to quantify the customers’ utility within a particular time period, is evaluated on a regular time basis. As the experience of every customer over time matters, it is therefore inappropriate to only aim at finding a good recommendation algorithm for final future customers as this may deteriorate the utility of current customers.
Hence, minimizing the simple regret is not a suitable objective in this case.
Instead, companies usually aim to boost the total utility scores of all customers throughout the entire hyper-parameter tuning process.
Another example of such applications is the development of combination therapies through clinical trials, where BO is often used to sequentially choose combinations of therapies for a series of patients, in order to improve the treatment efficacy \citep{SSWAF16,KDL20,TS21}.
In these applications, 
the focus is on the overall efficacy of combination therapies for all patients, and not only on the final patient.
%only aiming to find an effective combination therapy for future patients is unethical because it may result in ineffective treatments for the current patients. Therefore, the more appropriate objective for the doctors is 

\begin{figure}[h]
\includegraphics[scale=0.7]{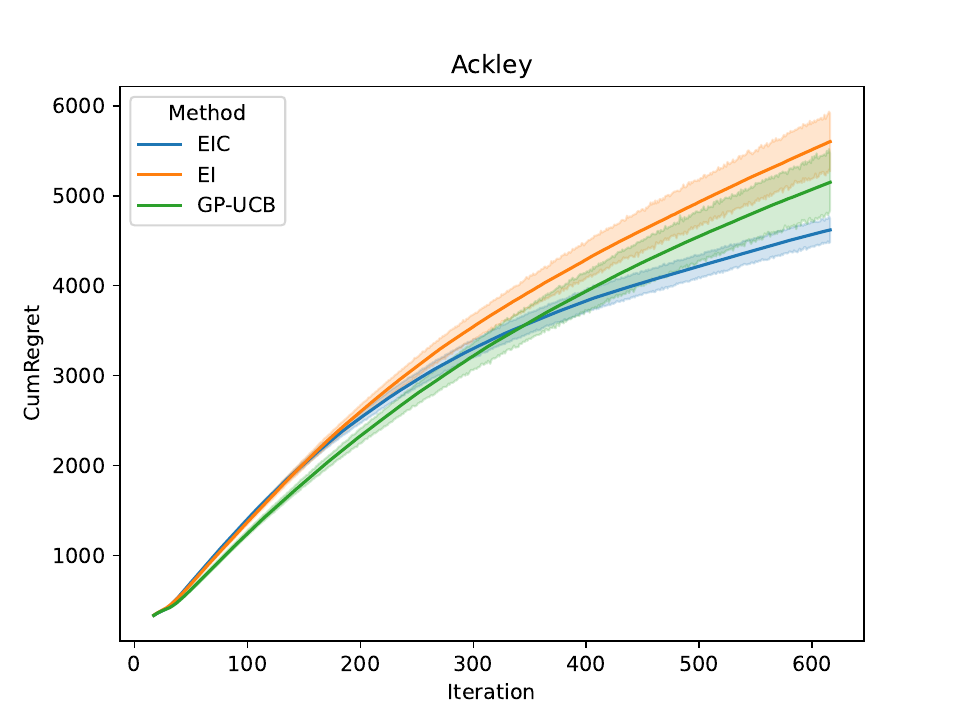}
\centering
\caption{\label{fig:1}
Cumulative regret of common BO methods and our proposed EIC algorithm on the Ackley test function. The solid line represents the cumulative regret (averaged over 100 independent runs) and the shaded area is the corresponding 95\% confidence region.
}
\end{figure}

%Introduce cumulative regret based on MAB literature
This objective of maximizing the overall performance throughout the entire experiment originated from the multi-armed bandit (MAB) literature, in which it is known as \emph{cumulative regret} minimization \citep{LR85,Agr19,LS20}.
Popular MAB approaches for this objective include upper confidence bound (UCB, \citealp{LR85,ACF02,CGMMS13,Lattimore18}) and Thompson sampling (TS, \citealp{Tho33,AG12,KKM13}), which are based on the frequentist and Bayesian perspectives, respectively.
Both approaches have been extended into the BO framework to derive the GP-UCB and GP-TS acquisition functions, and have been shown to perform well analytically under the cumulative regret setting \citep{Srinivas10,CG17,KKSP18,BSK19,VKP21}.
The acquisition function of EI has also been analyzed analytically under the cumulative regret evaluation metric \citep{WF14,Nguyen17}, and its existing performance bound  is shown to be not as tight compared to those of GP-UCB and GP-TS \citep{CG17}.
This theoretical gap in the cumulative regret of EI is also reflected in practice, which we illustrate using a numerical example in Figure \ref{fig:1}.
Here, we examine the performances of EI and GP-UCB on the commonly used Ackley function, which is a two-dimensional test function equipped with many local extrema (refer to Table \ref{tab:funcs} for more details).
The figure shows that GP-UCB (green curve), which is designed for the cumulative regret setting, incurs smaller cumulative regret than EI (orange curve).

%Bring up the question
In view of the above-mentioned theoretical and empirical gaps of the EI in cumulative regret minimization, the following question arises: \emph{can EI be adapted to achieve a tight upper bound on its cumulative regret which can then enable it to also perform well in applications focused on maximizing the overall performance?}
This is in fact an important open problem in BO because of the impressive real-world performances and wide adoption of EI.
To this end, we adapt the traditional EI algorithm to suit the objective of cumulative regret minimization, and propose the \emph{expected improvement-cost} (EIC) algorithm (Section \ref{section:EIC}).
We plot in Figure \ref{fig:1} the cumulative regret of EIC (blue curve), which shows that our proposed EIC achieving a smaller cumulative regret than traditional EI and performing comparably with GP-UCB.

%Briefly introduce EIC and contributions
The contributions of this paper are as follows.
\textbf{Firstly}, we propose the EIC algorithm (Section \ref{section:EIC}) for cumulative regret minimization.
EIC is designed to consciously balance the evaluation gains and losses in every iteration, so that a smaller cumulative regret is achieved.
The algorithm starts with a systematic and budget-dependent initial experimental design, which ensures that the global model fitting is reasonably good.
After the initialization, to choose a point to evaluate, EIC firstly calculates the evaluation cost of every point in the domain based on the expected loss (and the number of remaining iterations). 
This serves as a criterion to decide the worthiness of a point for evaluation, i.e., a point is worth evaluating only if its expected gain (i.e., the EI acquisition function value) is larger than its evaluation cost.
As a result, in every iteration, we sample the point that has the largest EI acquisition function value, provided that it is not lower than its evaluation cost.
If no point in the domain has an acquisition function value not lower than its evaluation cost, then the previous evaluated point with the best observed function value is sampled.
\textbf{Secondly}, we analyze the cumulative regret of our EIC algorithm and establish 
a high-probability finite-time regret upper bound.
Importantly, we show that EIC can achieve an upper bound of $O\big( \sqrt{N} \gamma_N (\log N)^{1/2} \big)$ under some mild regularity conditions, where $N$ is the total number of iterations and $\gamma_N$ is the maximum information gain (see Section \ref{subsection:acquisition}).
This regret upper bound is tighter than the bound of existing EI-based algorithms in the literature \citep{WF14,Nguyen17}.

%Remainder of the papar
The layout of the paper is as follows. 
In Section \ref{section:background}, we present some background information about BO and the GP model. We also give a brief review of the EI acquisition function and its related theoretical results. 
In Section \ref{section:EIC}, we describe our proposed EIC algorithm and explain some interesting insights into the development of EIC, including why the evaluation cost is applied.
In Section \ref{section:regret:analysis}, we establish a cumulative regret upper bound for EIC.
In Section \ref{section:experiments} we perform several numerical experiments to demonstrate the practical effectiveness of our EIC algorithm.
Finally, Section \ref{section:conclusion} concludes the paper.

\section{Problem Statement and Background}
\label{section:background}
Let $A^T$ denote the transpose of a vector or matrix $A$ and 
let $I_n$ denote the identity matrix of size $n$.
Let $\text{diag}(a_1, a_2,\dots, a_n)$ denote the diagonal matrix of size $n$ with the $(i,i)$ entry equal to $a_i$.
Let $\Vert\bx \Vert := \sqrt{x_1^2+\cdots+x_d^2}$ denote the Euclidean norm of vector $\bx = (x_1,\dots,x_d)^T$.
Let $a^+$ denote $\max(0,a)$ and
let $\phi(\cdot)$ and $\Phi(\cdot)$ denote the density and cumulative distribution functions of standard normal distribution, respectively.
Let $\mathcal{N}_k(\bmu,\Sigma)$ denote the $k$-dimensional multivariate normal distribution with mean vector $\bmu$ and covariance matrix $\Sigma$.
%Let $a \wedge b$ denote $\min(a,b)$,
%$\lfloor \cdot \rfloor$ (resp.~$\lceil \cdot \rceil$) denote the greatest (resp.~least) integer function and 
%Let $a^+$ denote $\max(0,a)$
Let $a_n \sim b_n$ if $\lim_{n \rightarrow \infty} (a_n/b_n)=1$,
$a_n = O(b_n)$ if $\limsup_{n \rightarrow \infty} |a_n/b_n| < \infty$ and $a_n = \Omega(b_n)$ if $\liminf_{n \rightarrow \infty} |a_n/b_n| > 0$.

BO aims to sequentially maximize an unknown objective function $f:D \to \mathbb{R}$, where $D = [0,1]^d\subseteq \mathbb{R}^d$.
In each iteration $n$, BO selects a point $\bx_n$ to evaluate, and receives a noisy observation $y_n = f(\bx_n) + \epsilon_n$.
As discussed in Section \ref{section:introduction}, we consider the objective of minimizing the cumulative regret after $N$ total iterations:
\begin{equation} \label{cumreg}
	R_N = \sum_{n=1}^{N} (f(\bx^*) - f(\bx_n)),
\end{equation}
where $\bx^* = \arg \max_{\bx \in D} f(\bx)$ denotes the location where $f$ attains the global maximum. 
In order to choose the sequential points $\bx_n$'s intelligently, BO usually models the objective function using a GP model.
We briefly introduce GPs in Section \ref{subsection:GP}, and refer the readers to \cite{WR06} and \cite{KHSS18} for more complete introduction to GPs.

\subsection{Gaussian Processes}
\label{subsection:GP}
The objective function in BO is typically modeled as a stationary Gaussian process.
We say a random function $f$ follows a prior distribution $\mathcal{GP}(\mu,\omega^2 k)$ with mean function $\mu: D \to \mathbb{R}$, covariance kernel $k: D \times D \to \mathbb{R}$ and signal variance $\omega^2$ if and only if the following condition holds:
For every finite set of points $X = (\bx_1,\dots,\bx_n)^T$, the values $\big( f(\bx_1),\dots,f(\bx_n) \big)^T \sim 
\mathcal{N}_n(\mu_X, \omega^2 K_{XX})$, which is an $n$-dimensional multivariate normal distribution with
\begin{align}
	 \mu_X & := \big( \mu(\bx_1), \dots, \mu(\bx_n) \big)^T,  \text{ and } \\
	 K_{XX} & := \begin{pmatrix}
		k(\bx_1,\bx_1) & k(\bx_1,\bx_2) &  \cdots & k(\bx_1,\bx_n)\\
		k(\bx_2,\bx_1) & k(\bx_2,\bx_2) &  \cdots & k(\bx_2,\bx_n) \\
		\vdots & \vdots &  \ddots & \vdots\\
		k(\bx_n,\bx_1) & k(\bx_n,\bx_2) &  \cdots & k(\bx_n,\bx_n)
	\end{pmatrix}.
\end{align}
\medskip
One useful result of the GP model is that the posterior distribution of $f$, conditioned on the sampled data, is still a GP.
Let $X = (\bx_1,\dots,\bx_n)^T$ and $Y = (y_1,\dots,y_n)^T$ denote the sampled points and the corresponding noisy observations up to iteration $n$.
{Suppose the noise terms $\epsilon_n$'s are independent and identically distributed normal random variables $\text{N}(0,\omega^2\lambda^2)$,
we have that}
\begin{equation} \label{gppost}
	f|(X,Y) \sim \mathcal{GP}(\mu_n,\omega^2 k_n), 
\end{equation}
with $\mu_n: D \to \mathbb{R}$ and $k_n: D \times D \to \mathbb{R}$ given by
\begin{eqnarray} \label{mun}
	\mu_n(\bx) & = &  k_{\bx X}(K_{XX} +  { \lambda^2} I_n)^{-1}(Y_n - \mu_X), \\ \label{kerneln}
	k_n(\bx,\bx^\prime) & = &k(\bx, \bx^\prime) - k_{\bx X}(K_{XX} + { \lambda^2} I_n)^{-1} k_{ X \bx^\prime},
\end{eqnarray}
where $k_{\bx X} = k_{X \bx }^T = \big( k(\bx_1,\bx),\dots,k(\bx_n,\bx) \big)$.
{
Note that $\omega^2$ and $\lambda^2$ are algorithm-specific parameters, which could possibly depend on $n$.}
Based on (\ref{gppost}), it can be deduced that 
%the pointwise posterior distribution follows Gaussian.
the predictive distribution at any point follows a Gaussian distribution.
That is, for any $\bx \in D$,
\begin{equation} \label{postpwd}
	f(\bx)|(X,Y) \sim \text{N}(\mu_n(\bx),\omega^2 \sigma_n^2(\bx)), \text{ with } \sigma_n^2(\bx) =	k_n(\bx,\bx).
\end{equation}

%explanation on GP model
The mean function and the covariance kernel serve as the prior for the GP model, and they reflect the initial belief about $f$.
Without loss of generality, the mean function $\mu(\bx)$ is usually set to be zero, indicating there is no prior knowledge on the global maxima location,
whereas the choice of covariance kernel is more varied.
In this paper, we consider positive definite covariance kernel functions that are isotropic and bounded, which is a quite general setting.
Typical examples that satisfy this condition include the squared exponential (SE) and the Mat\'ern kernel, which are perhaps the most popular covariance kernels in practice for Bayesian optimization \citep{SLA12,SSWAF16,Teck20}. 
Let $\mathbf{h} := (h_1, \dots, h_d)^T$ with $h_i > 0, \forall 1 \leq i \leq d$ be the length-scale parameter.
The Euclidean distance between $\bx$ and $\bx^\prime$, adjusted by the length-scale parameter $\mathbf{h}$, is given by
$$ \Vert \bx - \bx^\prime \Vert_{\mathbf{h}} := \sqrt{\Big(\frac{x_1-x_1^\prime}{h_1} \Big)^2 + \cdots + \Big(\frac{x_d-x_d^\prime}{h_d} \Big)^2}. $$
Define the SE covariance kernel as
\begin{equation} \label{KSE}
	k_{\text{SE}}(\bx,\bx^\prime) =  \exp \left(-\frac{\Vert \bx - \bx^\prime \Vert_{\bf h}^2}{2} \right),
\end{equation}	
 and the Mat\'ern kernel
\begin{equation}  \label{KMA}
	k_{\text{Mat\'ern}}(\bx,\bx^\prime)  =  \frac{2^{1-\nu}}{\Gamma(\nu)}\left(\sqrt{2\nu}  \Vert \bx - \bx^\prime \Vert_{\bf h} \right)^\nu B_\nu \left(\sqrt{2\nu} \Vert \bx - \bx^\prime \Vert_{\bf h} \right),
\end{equation}
where $\nu > 0$ is the smoothness parameter, and $B_\nu$ is the modified Bessel function.
The length-scale $\mathbf{h}$ and smoothness $\nu$ are the hyper-parameters of the GP model.
Standard methods can be applied to estimate them, such as the maximum likelihood method by \cite{SWN18} and the maximum a posteriori method by \cite{NY12}.

%The Euclidean distance between $\bx, \bx^\prime \in D$, adjusted by the length-scale parameter $\mathbf{h}$, is
%$$ \Vert \bx - \bx^\prime \Vert_{\mathbf{h}} := \sqrt{\Big(\frac{x_1-x_1^\prime}{h_1} \Big)^2 + \cdots + \Big(\frac{x_d-x_d^\prime}{h_d} \Big)^2}. $$
%and the Mat\'ern kernel
%\begin{equation}  \label{KMA}
%	k_{\text{Mat\'ern}}(\bx,\bx^\prime)  =  \frac{\tau^2}{2^{\nu-1}\Gamma(\nu)}\left(\sqrt{2\nu}  \Vert \bx - \bx^\prime \Vert_{\bf h} \right)^\nu B_\nu \left(\sqrt{2\nu} \Vert \bx - \bx^\prime \Vert_{\bf h} \right),
%\end{equation}
%$\nu > 0$ is the smoothness parameter, and $B_\nu$ is the modified Bessel function.

\subsection{Expected Improvement}
EI is one of the most popular acquisition functions under BO framework.
We review here EI as proposed by \cite{Mockus75} and \cite{JSW98} when the observations are evaluated without noise, i.e., $y_m = f(\bx_m)$.
The noisy version will be discussed in more detail in Section \ref{section:EIC}. 
Let $E_n[\cdot]$ denote the expectation with respect to the posterior distribution of $f(\bx)$ described in (\ref{postpwd}).
After $n$ iteration, the acquisition function of EI is defined as
\begin{equation} \label{EInf}
	\alpha_{n}^{EI}(\bx) := E_{n}( [f(\bx)- f(\bx_{n}^*)]^+ ),
\end{equation}
where $f(\bx_{n}^*) = \max_{1 \leq m \leq n} f(\bx_m)$ is the current best observation value.
This acquisition function~\eqref{EInf} quantifies the expected gain of sampling $\bx$ over the current best observation value,
% when the points $(\bx_{1},\dots,\bx_{n})$ have been evaluated,
and a point that maximizes (\ref{EInf}) will be sampled in the next iteration, i.e.,
$$ \bx_{n+1} = 	\argmax_{\bx \in D}\alpha_{n}^{EI}(\bx).$$
%EI is designed to maximize the rewards of every single iteration.
%Although this strategy is seemed to be myopic, many applications have shown that it performs well under the simple regret evaluation metric.
%Moreover, its computation cost is very low as the acquisition function (\ref{EInf}) has a closed form expression.
EI has been repeatedly shown to perform competitively under the simple regret evaluation metric.
Moreover, it is computationally convenient because the acquisition function (\ref{EInf}) has a closed-form expression.

\subsection{Related Works}
\label{section:related:works}

In the theoretical analyses of BO algorithms, the objective function is usually assumed to be fixed 
and belongs to the reproducing kernel Hilbert space (RKHS) associated with the covariance kernel in the GP model.
Based on the different focus or interest in the application, the analyses in literature can be classified into two categories: 
one which focuses on the analysis of the asymptotic properties of the obtained optimum or optimal solution (i.e., the simple regret), 
and the other which focuses on analysing the cumulative regret.
Intuitively, the analysis of asymptotic properties (i.e., simple regret) can be used to assess if the algorithm converges and how fast it converges to the true optimum or optimal solution, whereas the cumulative regret analysis informs on the performance of the algorithm throughout the entire experiment.

The theoretical analyses of the EI started from studying its asymptotic properties (i.e., simple regret).
Under the noise-free BO setting, \cite{VB10} showed that when the covariance kernel function is fixed and has finite smoothness, the EI converges to the global maxima almost surely for any objective function.
Subsequently, \cite{Bull11} proved that for any $d$-dimensional objective function,
the convergence rate of the EI is of the order $O(N^{-(\nu\wedge 1)/d})$ with $\nu > 0$ being the smoothness parameter of the covariance kernel.
The author also showed that a combination of EI and the $\epsilon$-greedy method converges at the near-optimal rate of $O(N^{-\nu/d})$.
Under the noisy setting, \cite{Ryzhov16} showed that a variant of EI can achieve the same convergence rate as the optimal computing budget allocation (OCBA) algorithm, which is an algorithm that has been shown to be asymptotically near-optimal.

\cite{Srinivas10} did the first cumulative regret analysis on BO algorithms.
They showed that {when the objective function belongs to the RKHS associated with certain covariance kernel}, the acquisition function of GP-UCB can achieve a finite-time regret upper bound of $R_N = O\Big(\gamma_N \sqrt{N} (\log N)^{3/2}\Big)$ with high probability.
%The term $\gamma_N$ is the maximal information gain between the sampled sequence $\{\bx_{n}\}_{n=1}^N$ and the underlying model. 
The term $\gamma_N$ is the maximum information gain about the objective function $f$ that can be obtained from any set of $N$ sampled points.
Its value is algorithm-independent and closely related to the effective dimension associated with the kernel.
%\cite{VKP21} showed that $\gamma_N = O\Big((\log N)^{d+1}\Big)$ for the SE kernel, and $\gamma_N = O\Big((\log N)^{d+1}\Big)$ for the Mat\'ern kernel.
%This subsequently implies that GP-UCB can achieve a regret upper bound of $R_N = O\Big(\sqrt{N}(\log N)^{d+5/2}\Big)$.
The regret bound of GP-UCB was later improved by \cite{CG17} to $R_N = O\Big(\gamma_N \sqrt{N}\Big)$.
In addition, they proposed the acquisition function of GP-TS and showed that it can attain a regret upper bound of $R_N = O\Big(\gamma_N \sqrt{N} (\log N)^{1/2} \Big)$.
As for the EI acquisition function under the noisy BO setting, \cite{WF14} replaced $f(\bx_n^*)$ in (\ref{EInf}) with the maximum of the posterior mean $\max_{\bx \in D} \mu_{n}(\bx)$, and showed that their algorithm achieves $R_N = O\Big(\gamma_N^{3/2} \sqrt{N} (\log N)\Big)$ with high probability.
\cite{Nguyen17} further considered replacing $f(\bx_n^*)$ with the current best observation $\max_{1 \leq m \leq n} y_m$,
and reduced the regret upper bound to $R_N = O\Big(\gamma_N \sqrt{N} (\log N)^{3/2}\Big)$.
However, their analysis depends on a pre-defined constant $\kappa > 0$ which is set to be small for good performances of EI, yet the constant term in their regret upper bound will explode quickly as $\kappa \to 0$.
The recent work of \cite{TGRV22} also performed theoretical analyses of EI, but they considered a regret definition that is different from our paper and these previous works.

\cite{SBC17} and its update \cite{SBC18} derived a universal regret lower bound that applies to all algorithms under the noisy BO setting.
The authors showed that for the SE and Mat\'ern kernel respectively, an expected regret of $E(R_N) = \Omega\Big(\sqrt{N}(\log N)^{d/4}\Big)$ and $E(R_N) = \Omega\Big(N^{\tfrac{\nu+d}{2\nu+d}}\Big)$ is unavoidable.
This strong result motivates us to examine whether the current regret upper bounds of EI algorithms are tight.
%They claimed that this result is stronger than the previous regret upper bound results as it is based on expectation rather than probability (where an expectation result can imply a high-probability result, but the converse may not be true).
%As for the maximum information gain,\cite{VKP21} showed that $\gamma_N = O\Big((\log N)^{d+1}\Big)$ for the SE kernel, and $\gamma_N = O\Big(N^{\tfrac{d}{2\nu+d}} (\log N)^{\tfrac{2\nu}{2\nu+d}} \Big)$ for the Mat\'ern kernel.
%This is likely due to the reliance on the maximal information gain in the current cumulative regret analysis, which is derived under the worst-case scenario over all possible sets of sampled points.
As shown from these previous results in the literature, for the EI, there exists a gap between the regret lower bound and the current regret upper bounds.
In order to reduce this gap, we introduce a novel quantity called the evaluation cost, and compare it with the EI acquisition function. 
At each iteration, a new point will be sampled only if its acquisition function exceeds its evaluation cost.
We name this new algorithm the expected improvement-cost (EIC) algorithm.
Moreover, we adopt an analysis framework from the BO literature and establish in Section \ref{section:regret:analysis} a finite-time regret upper bound of EIC as $O\Big(\sqrt{N}\gamma_N(\log N)^{1/2}\Big)$, which is tighter than the regret upper bound of \cite{WF14} and \cite{Nguyen17}. 

%which is tighter than the bound of traditional EI acquisition function \citep{Nguyen17} by a logarithmic factor of $1$.

\section{The Expected Improvement-Cost (EIC) algorithm}
\label{section:EIC}
In this section we describe in detail our expected improvement-cost (EIC) algorithm.
We first introduce in Section \ref{subsection:initialization} an experiment scheme for choosing the initial design points.
This scheme is to ensure that the initial GP model fit is good so that the posterior variance is not too large uniformly over the entire domain.
%It is also useful in providing a necessary condition to show that our EIC algorithm is near optimal in theory.
Next, we discuss in Section \ref{subsection:acquisition} how the incumbent function of EI should be selected in the setting of noisy BO, i.e., when the observations are corrupted by homogenous noises.
The EI has a built-in mechanism to trade-off between sampling points with high expected value (posterior mean) versus high uncertainty (posterior variance), which is a desirable property under the cumulative regret objective.
However, the EI acquisition function, as it was originally designed, only quantifies the potential upside of evaluating a point and overlooks the potential downside.
Under the objective of cumulative regret, it is important to also account for the potential downside of evaluating a point, because the value of the objective function in every iteration contributes to the performance measure (i.e., the cumulative regret).
Hence, this requires the algorithm to be more conservative than under the simple regret objective.
To this end, in Section \ref{subsection:eic:algorithm}, we propose to quantify this potential downside using an \emph{evaluation cost} function, which helps determine whether our EIC algorithm should evaluate existing good points or explore new points.

%Let $a^+$ denote $\max(a,0)$ and let $E_n(\cdot)$ denote the expectation with respect to the posterior distribution of objective function $f$ under GP model, conditioning on the sampled points $\bx_1,\dots,\bx_n$ with observations $y_1,\dots,y_n$.

\subsection{Initial Experiment Scheme}
\label{subsection:initialization}
A proper initial experiment scheme is important for Bayesian Optimization.
As highlighted in \cite{Bull11}, inappropriate selection of the initial design points may cause the algorithm to fail completely.
In this paper, we adopt an  initial experimental design  that  evenly spaces design points across the domain $D=[0,1]^d$.
Define the collection of the initial design points $B_M$, indexed by $M \in \mathbb{N}$, as 
\begin{equation} \label{idp}
	B_M = \Big\lbrace \bx \in D:  x_i = \frac{2k_i-1}{2M} \text{ for }  k_i  \in \lbrace 1, \dots, M \rbrace, i = 1, \dots,d \Big\rbrace.
\end{equation}
Following this scheme, given a user-specified parameter $M \in \mathbb{N}$, the interval $[0,1]$ along each dimension is divided into $M$ equal-sized segments.
As a result, the domain $D$ is partitioned into $M^d$ hyper-cubes, and the initial design points in $B_M$ are located at the centre of these hyper-cubes.
According to \eqref{idp}, the corresponding total number of initial design points is $n_0 = M^d$.

%Explain why adopt such a scheme 
The reason for adopting such an initial experiment scheme is to control the overall posterior variance of the GP model
by the following: For any point $\bx$, it can be shown that $\sigma_{n}^2(\bx)$ increases with the Euclidean distance $ \min_{1 \leq j \leq n} \Vert\bx - \bx_j \Vert$.
That is, the posterior variance has a positive correlation with the Euclidean distance to its nearest sampled point.
With the
initial experiment scheme \eqref{idp},
it can be seen that its nearest sampled point is the centre of hyper-cube to which it belongs.
As each hyper-cube has a diagonal length of $\sqrt{d}M^{-1}$, it is guaranteed that the distance is not more than $\tfrac{1}{2}\sqrt{d}M^{-1}$ for all points.
Therefore after our initial experiment design scheme, the posterior variance of the GP model is uniformly controlled by $M$, which is the number of segments along each dimension.

%choice of M
The next issue is to determine the value of $M$.
Setting $M$  too large will make the algorithm focus too much on global modelling, leaving little budget for optimization; 
On the other hand, setting  $M$ too small will result in a bad global model which can  substantially diminish the sample efficiency. Hence, we need to seek a balance in choosing $M$ so that the algorithm will converge fast enough without utilizing too much budget for the  initial design.
Inspired by the regret analysis in Section \ref{section:regret:analysis}, we find that a budget-related choice of 
\begin{equation} \label{initdesign}
	M = O(N^{1/2d})  [\text{hence } n_0 = O(N^{1/2})]
\end{equation}
 is desirable to achieve such a balance.

\subsection{Acquisition Function}
\label{subsection:acquisition}
The general form of EI acquisition function, after $n$ observations has been evaluated is 
\begin{equation} \label{EIacq}
	\alpha_{n}^{EI}(\bx) = E_n( [f(\bx)- \xi_n]^+) = (\mu_{n}(\bx)-\xi_n) \Phi\big( \tfrac{\mu_{n}(\bx)-\xi_n }{\omega_{n}\sigma_{n}(\bx)} \big) + \omega_{n}\sigma_{n}(\bx)\phi \big(\tfrac{\mu_{n}(\bx)-\xi_n }{\omega_{n}\sigma_{n}(\bx)} \big),
\end{equation}
where $\xi_n$ is the incumbent value and $\omega_{n}^2$ is the signal variance parameter.
In the noise-free BO setting, the incumbent $\xi_n$ is usually selected as the current best observed function value: $\xi_n = \max_{1 \leq i \leq n} f(\bx_i)$.
When the observations are corrupted by homogenous noises, the noise-free function values are not observable, 
and a natural replacement is the current best (noisy) observation: $\xi_n = \max_{1 \leq i \leq n} y_i$.
However, 
%when using current best observations as the incumbent, 
this can make the acquisition function very unstable due to the observation noises.
To address this issue, \cite{HANZ06} proposed to multiply (\ref{EIacq}) by a factor of $\sigma_n(\bx)/\sqrt{\sigma_n^2(\bx)+\lambda^2}$,
so that the acquisition function is discounted according to the parameter $\lambda^2$.
Another possible remedy is to rely on the GP model, for example \cite{BCF10} and \cite{WF14} suggested to use the best posterior mean $\max_{\bx \in D} \mu_{n}(\bx)$ as the incumbent,
and \cite{PGRC13} recommended a quantile-based incumbent
$\max_{1 \leq i \leq n} \mu_n(\bx_i) - \Phi^{-1}(\beta)\sigma_n(\bx_i)$
with quantile level parameter $\beta = 0.5$ or $0.9$.
In our EIC algorithm, we adopt the approach of \cite{BCF10} and \cite{WF14}, but adapt it by considering the current best posterior mean at only the observed locations as the incumbent. 
Specifically, we define the incumbent value after $n$ observations as:
\begin{equation} \label{incum}
	\xi_{n} = \max_{1 \leq i \leq n} \mu_n(\bx_{i}).
\end{equation}
This choice of incumbent is intuitive and allows us to derive our theoretical guarantee.
Intuitively, if the GP model fitting is good, then $\xi_{n}$ will be a good estimation of the current best function value with low variability.
Moreover, unlike the incumbent used in \cite{WF14}, (\ref{incum}) does not require optimizing the GP posterior mean over the entire domain and hence does not introduce excessive computational cost to the algorithm and uncertainty.

The signal variance parameter $\omega_{n}^2$ in the GP model is used to balance the exploration and exploitation of the EI algorithm.
When $\omega_{n}$ is large, EI tends to explore regions which have less points,
whereas a small $\omega_{n}$ makes EI prefer exploiting regions which are predicted to have good points based on the observations so far.
The signal variance parameter has been studied in many previous literature \citep{AG12,WF14,CG17,TGRV22},
and its choice is closely related to the maximum information gain, which is defined as follows.
Let $X_n = (\bx_1,\bx_2,\dots,\bx_n)^T$ denote the vector of sampled points and $F_n = (f(\bx_1),f(\bx_2),\dots,f(\bx_n))^T$ be the corresponding function values.
The maximum information gain at iteration $n$, denoted by $\gamma_{n}$, is
\begin{equation} \label{infogain}
	\gamma_{n} := \sup_{X_n \subset D} I(Y_n;F_n),
\end{equation}
where $I(Y_n;F_n)$ denotes the mutual information between $F_n$ and $Y_n = F_n+\boldsymbol{\epsilon}$ with $\boldsymbol{\epsilon} \sim \mathcal{N}_n( \mathbf{0}, \lambda^2 \omega_{n-1}^2 \boldsymbol{I}_n)$.
While $\gamma_{n}$ depends on the structure of covariance kernel and the domain knowledge,
its value is not tied to any specific algorithm \citep{VKP21}.
%To estimate the maximum information gain, we can use a greedy method by selecting a point with the largest posterior variance sequentially.	
Inspired by Theorem \ref{thm1}, we found that setting
\begin{equation} \label{omgn}
	\omega_{n} = O\big(\sqrt{\gamma_{n}+1+\log(1/\delta)}\big) \text{ for some } 0 < \delta < 1
\end{equation} 
guarantees the convergence of EIC.

\subsection{Evaluation Cost and the EIC Algorithm}
\label{subsection:eic:algorithm}
Under the traditional EI framework, a point with the largest acquisition function value (\ref{EIacq})
will be sampled in each iteration.
The acquisition function \eqref{EIacq} quantifies the expected gain over $\xi_n$ if the point $\bx$ is sampled in the next iteration, which is the potential upside of evaluating this point.
However, this traditional strategy does not take into account the potential downside of sampling $\bx$, which is important under cumulative regret considerations.
This is because the function value in every iteration contributes to the cumulative regret performance measure.
As a result, if the algorithm samples a point with a significantly inferior function value (i.e., with substantial downside), the cumulative regret will increase significantly.
%In order to better balance between the evaluation gains and losses hence to achieve a smaller cumulative regret, in EIC we consider an additional step of comparing acquisition function against an evaluation cost.
In this regard, in order to achieve a smaller cumulative regret, our EIC takes an additional step of comparing the acquisition function value against an \emph{evaluation cost}, which allows us to better balance between the evaluation gains and losses.
Specifically, we define the evaluation cost after $n$ observations as:
\begin{eqnarray} \label{evacost}
	L_n(\bx)  & := &  E_n([\xi_n-f(\bx)]^+)/(N-n) \\ \nonumber
	 & = & \big[(\xi_n -\mu_{n}(\bx)) \Phi\big( \tfrac{\xi_n-\mu_{n}(\bx)}{\omega_{n}\sigma_{n}(\bx)} \big) + \omega_{n}\sigma_{n}(\bx)\phi \big(\tfrac{\xi_n-\mu_{n}(\bx)}{\omega_{n}\sigma_{n}(\bx)} \big) \big] /(N-n).
\end{eqnarray}
The numerator here quantifies the expected loss of sampling $\bx$ if its function value is less than the incumbent $\xi_n$,
and the denominator is the number of remaining iterations.
%We shall explain why (\ref{evacost}) should be used as evaluation cost instead of the expected loss itself after describing our proposed algorithm.
We will first describe our complete EIC algorithm in the next paragraph, following which we will explain the intuitions behind the evaluation cost \eqref{evacost}.

\medskip
%\begin{algorithm}
%	\caption{Expected improvement-cost (EIC) algorithm}\label{EIalg}
%	\begin{algorithmic}
%		\Require $N$, $\mathcal{GP}(\mu,k)$, $n_0$, $b$.
%		\State Sample $n_0$ initial design points as described in (\ref{idp}). Each point is sampled with one replication.
%		\For{$n = n_0+1,\dots,N$}
%		\State Update the GP posterior model $\mathcal{GP}(\mu_n,k_n)$.
%		\If{ $\exists \bx \in D:\alpha_n^{EI}(\bx) \geq L_n(\bx)$}
%		\State Let $ B_n = \{\bx \in D: \alpha_n^{EI}(\bx) \geq L_n(\bx)\}$.
%		\State Sample the point satisfying $\argmax_{x: x\in B_n} \alpha_n^{EI}(\bx)$ with one replication.
%		\Else
%		\State Sample the point satisfying $\arg \max_{1 \leq i \leq M_n} U_i^n$ with one replication.
%		\EndIf
%		\EndFor
%	\end{algorithmic}
%\end{algorithm}
\begin{algorithm}
	\caption{Expected improvement-cost (EIC) algorithm}\label{EIalg}
	\begin{algorithmic}
		
		\Require $N$, $\mathcal{GP}(\mu,k)$, $n_0$, {$\xi_{n}$, $\omega_{n}^2$}.
		\State Sample $n_0$ initial design points as described in (\ref{idp}). Each point is sampled with one replication.
		\For{$n = n_0,\dots,N-1$}
		\State Update the GP posterior model $\mathcal{GP}(\mu_{n},k_{n})$ using the history of observations.
		\If{ $\exists \bx \in D:\alpha_{n}^{EI}(\bx) \geq L_{n}(\bx)$}
		\State Let $ B_n = \{\bx \in D: \alpha_{n}^{EI}(\bx) \geq L_{n}(\bx)\}$.
		\State Select the point $\bx_{n+1} = \argmax_{x: x\in B_n} \alpha_{n}^{EI}(\bx)$.
		\Else
		\State { Select the point $\bx_{n+1} =  \arg \max_{1 \leq i \leq n} \mu_n(\bx_i)$}.
		\EndIf
		\State Evaluate the selected point $\bx_{n+1}$ with one replication and observe $y_{n+1}$.
		\State Add the newly collected $(\bx_{n+1}, y_{n+1})$ to the history of observations.
		\EndFor
	\end{algorithmic}
\end{algorithm}

Algorithm \ref{EIalg} presents the pseudo-code of our EIC algorithm.
{It has 5 input parameters: the total budget $N$, the prior GP model, the total number of initial design points $n_0$ and the incumbent function $\xi_n$ and the signal variance parameter $\omega_{n}^2$}.
EIC starts with the initial experiment scheme (Section \ref{subsection:initialization}), where a total number of $n_0$ different points are sampled. 
These points are pre-determined by the initial design scheme (\ref{idp}) and each point is evaluated with one replication.
After the initial experiment, the subsequent points are sampled based on the acquisition function and evaluation cost.
%At each iteration the algorithm searches for a point that has the largest acquisition function, provided its value is not smaller than the evaluation cost of this point.
%If there exists such a point, then this point will be evaluated with one replication and we move on to the next iteration.
%If there is not point having acquisition function larger than the evaluation cost, 
%the algorithm will evaluate a previously sampled point which has the largest current upper confidence bound $U_i^n$ with one replication.
In each iteration $n$, if there exists at least one point $\bx$ which satisfies the condition of $\alpha_n^{EI}(\bx) \geq L_n(\bx)$ (i.e., its acquisition function value is not smaller than its evaluation cost), then we select the point with the largest acquisition function value among  all points  that satisfy this condition, and evaluate it with one replication.
If no point satisfies this condition, we select the previously sampled point with the largest {posterior mean} and add one more replication run to that point.

%explain evluation cost
\paragraph{Intuitions behind the Evaluation Cost.}
We use the case where the observations are noise-free to better illustrate the intuitions. 
In the noise-free case, the incumbent $\xi_n = \max_{1 \leq m \leq n} f(\bx_m)$ is the current best function value.
In iteration $n$, suppose we decide not to evaluate any new point but instead continue sampling at the current best point until the budget is exhausted, 
then the cumulative regret of the remaining iterations will be $(N-n)(f(\bx^*)- \xi_n)$.
On the other hand,
we can choose to sample a new point $\bx_{n+1}$ in the next iteration, and then stop evaluating any new point afterwards.
For this newly sampled point $\bx_{n+1}$, its function value $f(\bx_{n+1})$ may be larger or smaller than $\xi_n$.
In the first case where $f(\bx_{n+1}) \geq \xi_n$, an improvement is achieved and its expected value is given as $E_n([f(\bx_{n+1})-\xi_n]^{+})$.
In this case, we can choose to keep sampling at $\bx_{n+1}$ instead of $\argmax_{1 \leq m \leq n} f(\bx_m)$ in all the $(N-n)$ remaining iterations and this will provide a higher reward than with the continued sampling at $\xi_n$.
Hence, the total expected reduction of the cumulative regret (i.e., total expected gain) in all $(N-n)$ remaining iterations will be
\begin{equation*}
	(N-n)E_n([f(\bx_{n+1})-\xi_n]^{+}) = (N-n)\alpha_n^{EI}(\bx_{n+1}).
\end{equation*}
In the second case where $f(\bx_{n+1}) < \xi_n$, a loss will be incurred and the expected value will be $E_n([\xi_n-f(\bx_{n+1})]^+)$.
Unlike the improvement which can be exploited for all $(N-n)$ remaining iterations as mentioned above,
we only suffer this loss once (in iteration $n+1$) since we can switch back to the strategy of sampling $\argmax_{1 \leq m \leq n} f(\bx_{m})$ for all future iterations to avoid further this loss. This provides us a balance, and we see that $\bx_{n+1}$  is therefore worth sampling if and only if its \emph{total} expected gain is larger than its \emph{total} expected loss:
$$(N-n)\alpha_n^{EI}(\bx_{n+1}) \geq E_n([\xi_n-f(\bx_{n+1})]^+),$$
or equivalently: $\alpha_n^{EI}(\bx_{n+1})  \geq L_n(\bx_{n+1})$.

Another interesting insight is that the evaluation cost $L_n(\bx_{n+1})$ \eqref{evacost} increases with the iteration number $n$.
Therefore, our EIC algorithm is endowed with a built-in mechanism which allows it to more aggressively explore new points at the beginning and then gradually become more exploitative as the budget gets exhausted.
%Overall, in order to reduce the excessive exploration under the evaluation metric of cumulative regret, 
%EIC algorithm is more conservative than traditional EI due to the additional comparison of acquisition function with the evaluation cost,
%and the traditional EI can be regarded as a special case of EIC algorithm with zero evaluation cost.
Overall, our EIC algorithm is more conservative (in sampling new points) than the traditional EI as it evaluates the potential gains and losses by comparing the acquisition function with the evaluation costs. Further note that the traditional EI can be regarded as a special case of EIC algorithm with zero evaluation cost.
%Overall comments
%The definition of current best $\xi_n$ is optimistic with form $U_i$ 
%So it is harder for a new mean observation to beat this current best. It need to be better than 
%However, as we consider the total gain versus total cost, it has a natural mechanism that encourages more progressive sampling of new points at the start of the algorithm, and then becoming more conservative as the budget get exhausted.

\section{Regret Analysis of EIC}
\label{section:regret:analysis}
In this section we perform regret analysis of our proposed EIC algorithm.
We adopt the frequentist view by assuming that $f$ is an arbitrary function from the reproducing kernel Hilbert space (RKHS) associated with the covariance kernel in the GP model.
In Section \ref{sec:5.1}, we present a brief introduction to RKHS and its connections with GP models.
A complete overview of RKHS can be found in \cite{BT11}.
In Section \ref{sec:5.2}, we establish {a finite-time} cumulative regret upper bound for EIC.

\subsection{Reproducing Kernel Hilbert Space \label{sec:5.1}}

Let $\mathcal{X}$ be a non-empty set and $k(\cdot,\cdot): \mathcal{X} \times \mathcal{X} \to \mathbb{R}$ be a symmetric positive definite kernel.
Common examples of symmetric positive definite kernels are the SE and Mat\'ern  covariance kernel.
A Hilbert space $\mathcal{H}_{k}$ of functions on $\mathcal{X}$ equipped with an inner-product $\langle \cdot,\cdot \rangle_{\mathcal{H}_{k}}$ is called a \emph{reproducing kernel Hilbert space} (RKHS) with reproducing kernel $k$ if the following conditions are satisfied:
\begin{enumerate}
	\item For all $\bx \in \mathcal{X}$, we have $k(\cdot,\bx) \in \mathcal{H}_{k}$; 
	
	\item For all $\bx \in \mathcal{X}$ and all $f \in \mathcal{H}_{k}$, we have
	\begin{equation} \label{RPP}
		f(\bx) =  \langle f,k(\cdot,\bx)  \rangle_{\mathcal{H}_{k}}.
	\end{equation}
\end{enumerate}
%Based on the definition of RKHS, it can be deduced that a kernel function can be written as an inner-product product: For all $\bx,\bx^\prime \in\mathcal{X}$, 
Based on the definition of RKHS, it can be deduced that for all $\bx,\bx^\prime \in\mathcal{X}$, the kernel function $k(\bx,\bx^\prime)$ can be written as:
\begin{equation} \label{inrep}
	k(\bx,\bx^\prime) =  \langle k(\cdot, \bx),k(\cdot, \bx^\prime) \rangle_{\mathcal{H}_{k}}.
\end{equation}

%%justify why focus on RKHS
(\ref{inrep}) suggests that every RKHS defines a reproducing kernel function $k$ that is both symmetric and positive definite.
The other direction also holds as shown by the Moore-Aronszajn theorem \citep{Aronszajn50}, which states that given a positive definite kernel $k$, we can construct a unique RKHS of real-valued functions with $k$ as its reproducing kernel function.
That is, RKHSs and positive definite kernels are one-to-one: for every kernel $k$, there exits a unique associated RKHS, and vice versa.
When performing regret analysis under the frequentist view,
it is natural to assume that the objective functions $f$ belongs to the RKHS associated with the covariance kernel $k$ specified in the GP model.

\subsection{Regret Upper Bound of EIC \label{sec:5.2}}
\subsubsection{Regret Upper Bound}
%We state the near-optimal regret bound of EIC in Theorem \ref{thm1}, after describing the assumptions for which the theorem holds.
%Assume that
In order to derive the regret bound of EIC (Theorem \ref{thm1}), we make the following assumptions.
These are standard assumptions commonly used in the analysis of BO algorithms \citep{Srinivas10,CG17}.

\medskip
\noindent 
(A1) The objective function $f$ belongs to the RKHS $\mathcal{H}_{k}$ associated with the positive semi-definite kernel function $k(\cdot,\cdot)$.
Moreover, 
$k$ is isotropic (that is $k(\bx,\bx^\prime)$ depends only on $\Vert \bx -\bx^\prime \Vert$ and decreases with $\Vert \bx -\bx^\prime \Vert$) and $k(\bx,\bx) \leq 1, \forall \bx \in D$.

\medskip
\noindent (A2) The RKHS norm of the objective function $\Vert f \Vert_{\mathcal{H}_k} := \sqrt{\langle f,f \rangle_{\mathcal{H}_{k}}}$ satisfies
\begin{equation*} \label{rkhsb}
	\Vert f \Vert_{\mathcal{H}_k} \leq B \text{ for some } B > 0.
\end{equation*}
%The RKHS norm captures the smoothness of $f$: as $\Vert f \Vert_{\mathcal{H}_k}$ increases, $f$ gets less smooth and the optimization problem becomes harder.

\medskip
\noindent
(A3) The noise sequence $\lbrace \epsilon_n\rbrace_{n\geq 1}$ is conditionally $R$-sub-Gaussian for a fixed constant $R >0$:
\begin{equation*}
	\forall n \geq 0, \forall \lambda \in \mathbb{R}, E[e^{\lambda \epsilon_n}|\mathcal{F}_{n-1}] \leq \tfrac{\lambda^2 R^2}{2},
\end{equation*}
where $\mathcal{F}_{n-1}$ is the $\sigma$-algebra generated by random variables $\lbrace \bx_1,\dots, \bx_{n-1}, \epsilon_1 ,\dots, \epsilon_{n-1}, \bx_{n} \rbrace$.

Note that these assumptions do not directly reflect the design of our EIC algorithm:
Although the objective function $f$ is fixed, unknown, and a member of $\mathcal{H}_{k}$, 
and the noise random variables $\epsilon_n$ are conditionally $R$-sub-Gaussian martingale difference sequence,
we still run EIC algorithm under the Gaussian process model as described in Section 2.1.
In general, this represents a mis-specified prior and noise model, which is known as the agnostic setting in \cite{Srinivas10}.

%\medskip
%\noindent (A3) The total number of iterations for the initial experiment in EIC algorithm satisfies
%\begin{equation} \label{nnot}
%	n_0 := C_0 \sqrt{N}[\log N]^{d/4} \text{ for some } C_0 >0.
%\end{equation}

%\begin{equation} \label{nnot}
%	n_0 := C_0 \sqrt{N}[\log N]^{-1/2} \text{ for some } C_0 >0.
%\end{equation}

%\medskip
%\noindent (A4) Let $ X = (\bx_1, \dots, \bx_n)$ be the sampled points up to iteration $n$,
%\begin{equation} \label{pvarmun}
%	\text{Var}[\mu_n(\bx)] = \sigma^2 k_{\bx X} (K_{XX} + \sigma^2 I_n)^{-1} (K_{XX} + \sigma^2 I_n)^{-1} k_{ X\bx}
%\end{equation}

%In Theorem \ref{thm1} below, for technical reasons we consider the following modification of EIC algorithm:
%\begin{itemize}
%	\item Define the evaluation cost as $L_n(\bx) = E_n([\xi_n-f(\bx)]^+)/N$.
%\end{itemize}
%Intuitively, for an algorithm to achieve near-optimal cumulative regret bound, the algorithm should $o(N)$

\medskip
\begin{thm} \label{thm1}
	Assume {\rm (A1)}--{\rm (A3)}.
	Let $0 < \delta <1$. 
	With probability at least $1-\delta$, running EIC algorithm under a GP model with prior mean function $\mu(\bx) \equiv 0$, and with parameters (\ref{initdesign}), (\ref{incum}), (\ref{omgn}), $\lambda^2 = 1 + 2/N$ achieves
	\begin{equation} \label{EIreg}
		R_N = O(\sqrt{N}\gamma_N (\log N)^{1/2}).
	\end{equation}
\end{thm}

\begin{remark}
Under the noisy BO setting, \cite{WF14} proved that the EI algorithm can achieve a regret upper bound of $R_N = O\big(\sqrt{N} \gamma_N^{3/2} (\log N)\big)$.
Thereafter, \cite{Nguyen17} derived an improved regret upper bound of EI.
However, to the best of our knowledge, there is an issue in their proofs as highlighted in \cite{TGRV22}.
%(Specifically, they generalized (\ref{postvarsum}) in Lemma~5 as 
%$\sum_{n = 0}^{N-1} \sigma_{n}^2(\bx) \leq 2 \lambda^2\gamma_{N}, \forall \bx \in D$,  
%which is not true.)
Compared with \cite{WF14}, the regret upper bound in Theorem \ref{thm1} is much tighter,
which improves upon traditional EI by a factor of $\sqrt{\gamma_N \log N}$.
This significant improvement is achievable due to the evaluation cost feature in our EIC algorithm.

High-probability regret upper bounds which hold for every finite budget $N$ have been established for a number of BO algorithms, such as GP-UCB and GP-TS.
Under the frequentist setting, 
the tightest regret upper bound for vanilla GP-UCB and GP-TS are $R_N = O\big(\sqrt{N} \gamma_N \big)$ and $R_N = O\big(\sqrt{N} \gamma_N (\log N)^{1/2}\big)$ respectively as shown by \cite{CG17}.
To the best of our knowledge, our Theorem \ref{thm1} is the first regret bound for EI-based algorithms which matches the regret bound of GP-UCB and GP-TS.
\end{remark}

\cite{VKP21} provided general upper bounds of the maximum information gain, based on the decay rate of the eigenvalues of the covariance kernel function.
Using their results for the SE and Mat\'ern kernels, we can derive the following corollary.
\begin{cor}
	Let $0 < \delta < 1$. With probability at least $1-\delta$, the regret of EIC algorithm satisfies
	\begin{eqnarray*}
		R_N & = & O\Big(\sqrt{N}(\log N)^{d+\tfrac{3}{2}} \Big) \text{ for the SE kernel,}\\
		R_N & = & O\Big(N^{\tfrac{2\nu+3d}{4\nu+2d}}(\log N)^{\tfrac{6\nu+d}{4\nu+2d}} \Big)\text{ for the Mat\'ern kernel.} 
	\end{eqnarray*}
\end{cor}
Comparing with \cite{SBC18}, the regret upper bound of EIC for the SE covariance kernel is tight up to logarithmic factors.
%\cite{SBC17} and the update \cite{SBC18} showed that for any objective function belonging to the RKHS of the SE and Mat\'ern kernel, an expected regret lower bound of $E(R_N) = \Omega\big(\sqrt{N} (\log N)^{d/4}\big)$ and $E(R_N) = \Omega\big(N^{\tfrac{\nu+d}{2\nu+d}}\big)$ is unavoidable respectively. 
%This regret lower bound is a universal result and hence applies to all BO algorithms.
%we have $\gamma_{N} = O\big( (\log N)^{d+1} \big)$ for the SE kernel and  $\gamma_{N} = O\big( N^{\tfrac{d}{2 \nu +d}}(\log N)^{\tfrac{2\nu}{2\nu +d}} \big)$ for the Matern kernel.

%\medskip
%\begin{remark}
%   { 
%
%Theorem \ref{thm1} shows that EIC has a regret upper bound of $O\big(\sqrt{N} (\log N)^{d/2}\big)$,
%which has a slight gap of order $(\log N)^{d/4}$ with the universal regret lower bound,
%but is better than other upper bound results currently shown in the literature.
%}
%\end{remark}

%\begin{remark}
%	It is worth noting that our proof methodology may be extended to other covariance kernels, in addition to the SE kernel considered in this paper. The challenge arises from the control of the posterior variance (see Lemma \ref{lm2} in the Appendix). For example, for the Mat\'ern kernel, its posterior variance depends on a different form of filled distance compared to the SE kernel. 
%	Hence, extension to other kernels requires generalizaiton of our Lemma \ref{lm2} which is currently specific to the SE kernel.
%	%	Hence Lemma 2 is a kernel-specific posterior variance bound and needs to be further generalized.
%\end{remark}

%\subsubsection{Sketch of Proof}

\subsubsection{Proof of Theorem \ref{thm1}}

We preface the proof of Theorem 1 with Lemmas 1--7. 
These lemmas are proved in the appendix.
% In order to give a high probability regret upper bound of EIC algorithm, we need a few lemmas.
The first lemma is borrowed from \cite{CG17}, which provides a uniform confidence interval for the objective function based on maximum information gain.
\begin{lem}[Theorem 2 of \cite{CG17}] \label{lmCG}
	Let $\lbrace \bx_n \rbrace_{n \geq 1}$ be a discrete time stochastic process which is measurable with respect to the $\sigma$-algebra $\lbrace \mathcal{F}_{n-1} \rbrace_{n\geq 1}$.
	Let $\lbrace \epsilon_n \rbrace_{n \geq 1}$ be a real-valued stochastic process such that is (a) $\mathcal{F}_{n}$-measurable, and (b) $R$-sub-Gaussian conditionally on $\mathcal{F}_{n-1}$ for some $R \geq 0$.
	Let $f$ be a member	of the RKHS of real-valued functions with RKHS norm bounded by $B$.
	Define $\beta_n = B + R\sqrt{2(\gamma_{n}+1+\log(1/\delta))}$. 
	Let $0<\delta<1$, we have
	\begin{equation*}
		P(\forall 0 \leq n \leq N-1, \forall \bx \in D, |f(\bx) - \mu_{n}(\bx)| \leq \beta_n \sigma_{n}(\bx)) \geq 1- \delta ,
	\end{equation*}
	where $\mu_{n}(\bx), \sigma_{n}^2(\bx)$ are posterior mean and variance kernel as defined in (\ref{postpwd}), with $\lambda^2 = 1+\tfrac{2}{N}$.
\end{lem}

\medskip
	The second lemma plays a key role in the proof of Theorem 1.
	It states that if the acquisition function of point a $\bx$ is larger than its evaluation cost, then the quantity $z_{n}(\bx) =  \tfrac{\mu_{n}(\bx)-\xi_n}{\sigma_{n}(\bx)}$ must be lower bounded.
	This explains why EIC can achieve a smaller cumulative regret than traditional EI.
	Lemma \ref{lm2} is a crucial technical contribution of the paper.
	Moreover, the lower bound of $z_n(\bx)$ in (\ref{EIC_bound}) is asymptotically tight and hence cannot be further improved.
	\begin{lem} \label{lm2}
		Let $z_{n}(\bx) :=  \tfrac{\mu_{n}(\bx)-\xi_n}{\sigma_{n}(\bx)}$. For $1 \leq n \leq N-1$,
		\begin{equation} \label{EIC_bound}
			\alpha_n^{EI}(\bx) \geq L_n(\bx) \text{ only if }   z_n(\bx) \geq -\omega_n \sqrt{2 \log (N-n)}.
		\end{equation} 
	\end{lem}

\medskip
	The third lemma provides upper and lower bounds on the EI acquisition function, which is based on \cite{Bull11} and \cite{WF14}.
	\begin{lem}  \label{lm3}
		Let $I_n(\bx) = \max \lbrace 0, f(\bx)-\xi_n \rbrace$ and let $0 < \delta < 1$. With probability at least $1-\delta$,
		\begin{equation*}
			\max \Big\lbrace I_{n}(\bx) - \beta_{n} \sigma_{n}(\bx), \tfrac{h(-\beta_{n}/\omega_{n})}{h(\beta_{n}/\omega_{n})} I_{n}(\bx) \Big\rbrace\leq \alpha_{n}^{EI}(\bx) \leq I_{n}(\bx) + (\beta_{n} + \omega_{n}) \sigma_{n}(\bx).
		\end{equation*}
	\end{lem}

\medskip
	It can be seen from Lemma \ref{lm3} that the upper bound of acquisition function depends on $I_n(\bx)$ and $\sigma_n(\bx)$. 
	In Theorem \ref{thm1}, we further bound these two quantities based on Lemma \ref{lm4} and \ref{lm5}.
	\begin{lem} \label{lm4}
		Let $I_n(\bx) = \max \lbrace 0, f(\bx)-\xi_n \rbrace$ with incumbent $\xi_n := \max_{1 \leq i \leq n} \mu_{n}(\bx_{i})$. 	Let $0 < \delta < 1$, we have
		\begin{equation*}
			P\Big( \sum_{n = 1}^{N-1} I_n(\bx_{n+1}) \leq \sqrt{2}B + \beta_{N} \sum_{n = 0}^{N-1} \sigma_{n}(\bx_{n+1 }) \Big) \geq 1- \delta.
		\end{equation*}
	\end{lem}

	\begin{lem} \label{lm5}
		For any $N \geq 1$, the points $\bx_1, \bx_2, \dots, \bx_N$ selected by any algorithm satisfy
		\begin{equation*}
			 \sum_{n = 0}^{N-1} \sigma_{n}^2(\bx_{n+1}) \leq 2 \lambda^2\gamma_{N}.
		\end{equation*}
	\end{lem}

\medskip
Lemma \ref{lem2B} bounds the function value difference for any two points within the domain based on RKHS norm.
\begin{lem} \label{lem2B}
	Assume {\rm (A1)} and {\rm (A2)}. $\forall \bx, \bx^\prime \in D$, we have $f(\bx) - f(\bx^\prime) \leq \sqrt{2} B$.
\end{lem}

\medskip
The last lemma shows that the posterior variance kernel is monotonically decreasing in the number of iterations.
\begin{lem} \label{lemsubsigma}
	$\forall \bx \in D$ and $n \geq 1$, we have
	\begin{equation}
		\sigma_n^2(\bx) = \sigma_{n-1}^2(\bx) - \frac{k_{n-1}^2(\bx,\bx_n)}{ \sigma_{n-1}^2(\bx_n) + \lambda^2 },
	\end{equation}
	where $k_{n}(\cdot,\cdot)$ are posterior covariance kernel as defined in (\ref{kerneln}).
	Hence $\sigma_n^2(\bx)$ is monotonically decreasing in $n$.
\end{lem}

%   {
%{\sc Proof sketch of Theorem \ref{thm1}}.
%Consider the decomposition of instantaneous regret into two parts:
%\begin{equation}
%	r_n := f(\bx^*) - f(\bx_{n+1}) =  \underbrace{f(\bx^*) - \xi_n}_{\text{term 1}} +   \underbrace{ \xi_n - f(\bx_{n+1})}_{\text{term 2}}.
%\end{equation}
%Using Lemma \ref{lmCG} and \ref{lm3}, term 1 can be bounded as
%\begin{equation} \label{term1upbound}
%	f(\bx^*) - \xi_n \leq c[ I_n(\bx_{n+1}) +(\beta_{n} + \omega_n) \sigma_{n}(\bx_{n+1}) + ] 
%\end{equation}
%for some constant $c > 0$. \\
%Bounding the second term requires the evaluation cost feature of EIC. By Lemma 1, 2 and 3, we have
%\begin{equation} \label{term2upbound}
%	\xi_n - f(\bx_n) \leq [\omega_n \sqrt{2 \log (N-n)} + \beta_{n}] \sigma_{n-1}(\bx_{n}).
%\end{equation}
%Finally, since $R_N = \sum_{n=1}^N r_n$, summing the instantaneous regret and applying Lemma 4 and 5 gives us the desired result.
%}

\bigskip
{
	\textsc{PROOF OF THEOREM \ref{thm1}}. 
		Let $z_n(\bx) = \tfrac{\mu_{n}(\bx)-\xi_n }{\sigma_{n}(\bx)}$ and let $h(x) = x \Phi(x) +\phi(x)$. 
	Express the acquisition function and the evaluation cost function as:
	\begin{eqnarray}  \label{alphanh}
		\alpha_{n}^{EI}(\bx) &  = & \omega_n\sigma_{n}(\bx) h\big( \tfrac{z_n(\bx) }{\omega_n} \big), \\ \label{Lnh}
		% \big[ \tfrac{z_n(\bx) }{\omega_n}\Phi\big( \tfrac{z_n(\bx) }{\omega_n} \big) + \phi\big( \tfrac{z_n(\bx) }{\omega_n} \big) \big] 
		L_n(\bx)  &	=  &\omega_n\sigma_{n}(\bx) h\big(-\tfrac{z_n(\bx) }{\omega_n} \big) /(N-n).
		%\big[-\tfrac{z_n(\bx)}{\omega_n}\Phi\big(-\tfrac{z_n(\bx)}{\omega_n} \big) + \phi\big(-\tfrac{z_n(\bx)}{\omega_n} \big) \big] 
	\end{eqnarray}
	Let $r_n = f(\bx^*) - f(\bx_{n+1})$, then the cumulative regret can be written as
	\begin{equation*}
		R_N = \sum_{n=0}^{n_0-1} r_n + \sum_{n= n_0 }^{N-1} r_n.
	\end{equation*}
	By (\ref{initdesign}) and Lemma \ref{lem2B},
	\begin{equation*}
		\sum_{n=0}^{n_0-1} r_n \leq \sqrt{2}B n_0 = O(\sqrt{N}).
	\end{equation*}
	Hence to show Theorem \ref{thm1}, it suffices to show that
	\begin{equation} \label{keyarg}
	 \sum_{n = n_0 }^{N-1} r_n = O\big(\sqrt{N} \gamma_{N} (\log N)^{1/2} \big).
	\end{equation}
	For $n \geq n_0$, we break the regret at iteration $n$ as 
	\begin{equation*}
		r_n = f(\bx^*) - f(\bx_{n+1}) =  [f(\bx^*) - \xi_{n}] + [\xi_{n} - f(\bx_{n+1})] =  [\textrm{I}] + [\textrm{II}].
	\end{equation*}
	\medskip
	\textbf{Bounding term [\textrm{I}]}. Since $\omega_{n} = O \big(\sqrt{\gamma_{n}+1+\log(1/\delta)} \big)$, we have $\beta_n/\omega_n \leq C$ for some $C > 0$.	
	Hence by Lemma \ref{lm3}, with probability at least $1-\delta$,
	\begin{equation} \label{insregterm1}
	[\textrm{I}] \leq I_{n}(x^*)  \leq \tfrac{h(\beta_n/\omega_n)}{h(-\beta_n/\omega_n)}   \alpha_{n}^{EI}(\bx^*)\leq C^\prime \alpha_{n}^{EI}(\bx^*) \text{ for some } C^\prime > 0.
	\end{equation}
	To bound $\alpha_{n}^{EI}(\bx^*)$, we consider two cases:
	\begin{itemize}
		\item $\alpha_n^{EI}(\bx^*) \geq L_n(\bx^*)$ at iteration $n$. By Lemma 3,
		\begin{equation}  \label{term1case1}
			\alpha_{n}^{EI}(\bx^*)
			\leq   \alpha_{n}^{EI}(\bx_{n+1}) 
			\leq   I_n(\bx_{n+1}) + (\beta_n + \omega_n) \sigma_{n}(\bx_{n+1}), 
		\end{equation}
		where the first inequality follows because $\bx_{n+1}$ has the largest acquisition function value.
		\item $\alpha_{n}^{EI}(\bx^*) < L_n(\bx^*) $ at iteration $n$.
		By (\ref{alphanh})--(\ref{Lnh}), this implies that
		\begin{equation} \label{zless0}
			h\big(\tfrac{z_n(\bx^*)}{\omega_n}\big) < h\big(-\tfrac{z_n(\bx^*)}{\omega_n}\big)/(N-n) \leq h\big(-\tfrac{z_n(\bx^*)}{\omega_n}\big).
		\end{equation}
		Since $h(z) = z +h(-z)$, we can deduce from (\ref{zless0}) that $z_n(\bx^*) \leq 0$ when $\alpha_{n}^{EI}(\bx^*) < L_n(\bx^*) $.
		Let $\bx_{n}^* = \argmax_{1 \leq i \leq n} \mu_{n}(\bx_{i})$.
		We have
		\begin{eqnarray} \label{minusznxstar}
			h\big(-\tfrac{z_n(\bx^*)}{\omega_n}\big) &  \leq &\frac{\xi_n - \mu_{n}(\bx^*)}{\omega_n \sigma_{n}(\bx^*)} + 1 \\ \nonumber
			& \leq  &\frac{f(\bx_{n}^*) + \beta_n \sigma_{n}(\bx_{n}^*) - f(\bx^*) +  \beta_n \sigma_{n}(\bx^*)}{\omega_n \sigma_{n}(\bx^*)} + 1 \\ \nonumber
			& \leq & \frac{\sqrt{2}B + 2\beta_n}{\omega_n \sigma_{n}(\bx^*)} + 1,
		\end{eqnarray}
		where the first inequality follows from $h(z) \leq z + 1$ when $z \geq 0$, the second inequality follows from Lemma 1 and the last inequality follows from Lemma \ref{lem2B}, Lemma \ref{lemsubsigma} and $\sigma_{0}(\bx) \leq 1$. 
		Substitute (\ref{minusznxstar}) back into (\ref{Lnh}) gives us
		\begin{equation} \label{term1case2}
			\alpha_{n}^{EI}(\bx^*) < L_n(\bx^*)  \leq (\sqrt{2}B + 2\beta_n + \omega_n)/(N-n).
		\end{equation}
	\end{itemize}
	 By (\ref{insregterm1}), (\ref{term1case1}) and (\ref{term1case2}), we can deduce that
	 \begin{equation} \label{term1ub}
	 	[\textrm{I}] \leq C^\prime [ I_n(\bx_{n+1}) + (\beta_n + \omega_n) \sigma_{n}(\bx_{n+1}) +  (\sqrt{2}B + 2\beta_n + \omega_n)/(N-n)]. \\
	 \end{equation}
 %	\begin{eqnarray} \label{insregterm1}
	%		[\textrm{I}] & \leq & I_{n}(x^*) \\ \nonumber
	%		& \leq & \tfrac{h(\beta_n/\omega_n)}{h(-\beta_n/\omega_n)} \alpha_{n}^{EI}(\bx^*) \\ \nonumber
	%		& \leq & \tfrac{h(\beta_n/\omega_n)}{h(-\beta_n/\omega_n)} \alpha_{n}^{EI}(\bx_n) \\ \nonumber
	%		& \leq & \tfrac{h(\beta_n/\omega_n)}{h(-\beta_n/\omega_n)} [ I_n(\bx_n) + (\beta_n + \omega_n) \sigma_{n-1}(\bx_n)] \\ \nonumber
	%		& \leq & c I_n(\bx_n) + c(\beta_N + \omega_N) \sigma_{n-1}(\bx_n).
	%	\end{eqnarray}
%	The second and forth inequality of (\ref{insregterm1}) follows from Lemma 2, and the third inequality follows from the definition of acquisition function. \\
	\medskip
	\textbf{Bounding term [\textrm{II}]}. Suppose there exist an $\bx \in D$ at iteration $n$ such that $\alpha_n^{EI}(\bx) \geq L_n(\bx)$, then by Lemma 1 and Lemma 3, with probability at least $1-\delta$,
	\begin{eqnarray} \label{insregterm2}
		[\textrm{II}] & = &  \xi_n - f(\bx_{n+1}) \\  \nonumber
		& = & \big(-\tfrac{\mu_{n}(\bx_{n+1})-\xi_n}{\sigma_{n}(\bx_{n+1})} + \tfrac{\mu_{n}(\bx_{n+1})-f(\bx_{n+1})}{\sigma_{n}(\bx_{n+1})} \big)  \sigma_{n}(\bx_{n+1}) \\ \nonumber
		& \leq &\big[ \omega_n \sqrt{2 \log (N-n)}+\beta_n \big] \sigma_{n}(\bx_{n+1}).
	\end{eqnarray}
	On the other hand, suppose there exists no such point, then $\bx_{n+1} = \argmax_{1 \leq i \leq n} \mu_{n}(\bx_i) $. 
	Hence by Lemma 1, $\xi_n - f(\bx_{n+1}) = \mu_n(\bx_{n+1}) - f(\bx_{n+1})\leq \beta_n \sigma_{n}(\bx_{n+1})$ and (\ref{insregterm2}) still holds.\\
	\medskip
	Combining (\ref{term1ub}), (\ref{insregterm2}) and Lemma 5:
	\begin{eqnarray} \label{RNfinal}
	\sum_{n = n_0}^{N-1} r_n  & \leq & \sum_{n = n_0}^{N-1}  \Big\lbrace C^\prime I_n(\bx_{n+1}) + C^\prime (\sqrt{2}B + 2\beta_{n} + \omega_n)  (N-n)^{-1} \\ \nonumber
	& \quad & \quad+ \big[(C^\prime+\sqrt{2 \log (N-n)} ) \omega_n  + (C^\prime+1) \beta_n \big] \sigma_{n}(\bx_{n+1})  \Big\rbrace \\ \nonumber
	& \leq &  C^\prime \sum_{n=1}^{N-1} I_n(\bx_{n+1}) + C^\prime (\sqrt{2}B + 2\beta_{N} + \omega_N) \sum_{n = 1}^{N-1} (N-n)^{-1} \\ \nonumber
	& \quad & \quad+ \big[(C^\prime+\sqrt{2 \log N} ) \omega_N  + (C^\prime+1) \beta_N \big] \sum_{n = 0}^{N-1} \sigma_{n}(\bx_{n+1}) \\ \nonumber
	& \leq &  C^\prime \sqrt{2}  B + C^\prime (\sqrt{2}B + 2\beta_{N} + \omega_N) (\log N +1) \\ \nonumber
	& \quad & \quad+ \big[(C^\prime+\sqrt{2 \log N} ) \omega_N  + (2C^\prime+1) \beta_N \big] \sum_{n = 0}^{N-1} \sigma_{n}(\bx_{n+1}),
	\end{eqnarray}
	where the last inequality follows from Lemma \ref{lm4} and $\sum_{i=1}^{N} (N-n)^{-1} \leq  \log N + 1$.
	By the Cauchy-Schwarz inequality, $\sum_{n = 0}^{N-1} \sigma_{n}(\bx_{n+1}) \leq \sqrt{N\sum_{n = 0}^{N-1} \sigma_{n}^2(\bx_{n+1})}$.
	Since $\beta_{N} = O(\sqrt{\gamma_{N}})$, $\omega_{N} = O(\sqrt{\gamma_{N}})$ and $\lambda^2 = 1 +\tfrac{2}{N}$,
	applying Lemma \ref{lm5} to (\ref{RNfinal}) gives us (\ref{keyarg}).
	Theorem \ref{thm1} follows.
	$\wbox$
}

\section{Experiments}
\label{section:experiments}

\begin{figure}[!t]
	\centering
	\begin{subfigure}{0.5\linewidth}
		\includegraphics[width=\linewidth]{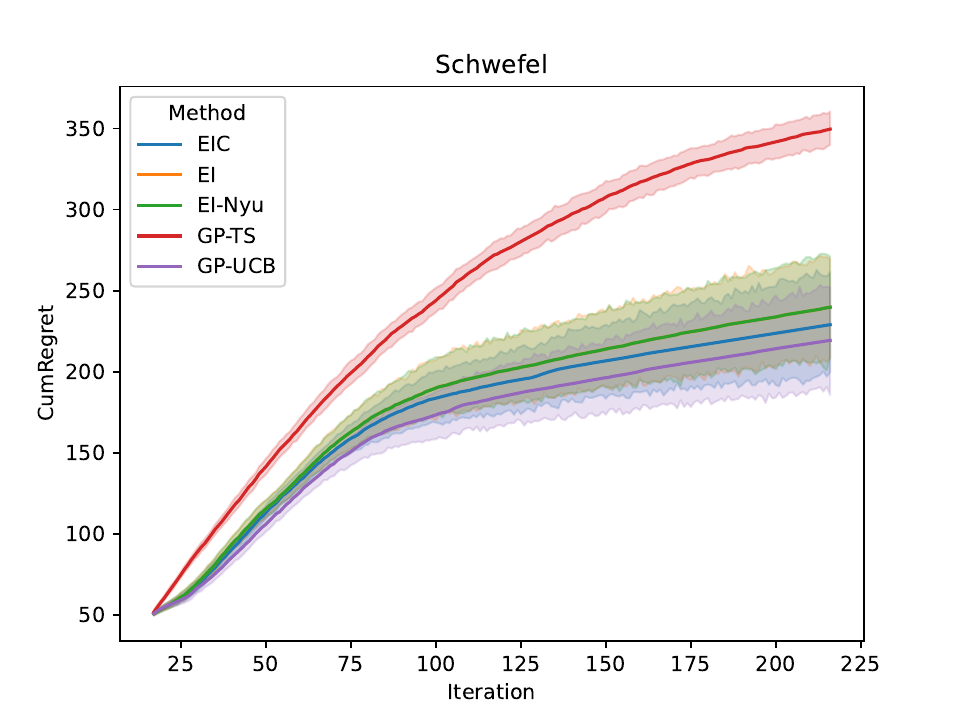} 
		\caption{Schwefel-2}
		\label{fig:2a}
	\end{subfigure}\hfill
	\begin{subfigure}{0.5\linewidth}
		\includegraphics[width=\linewidth]{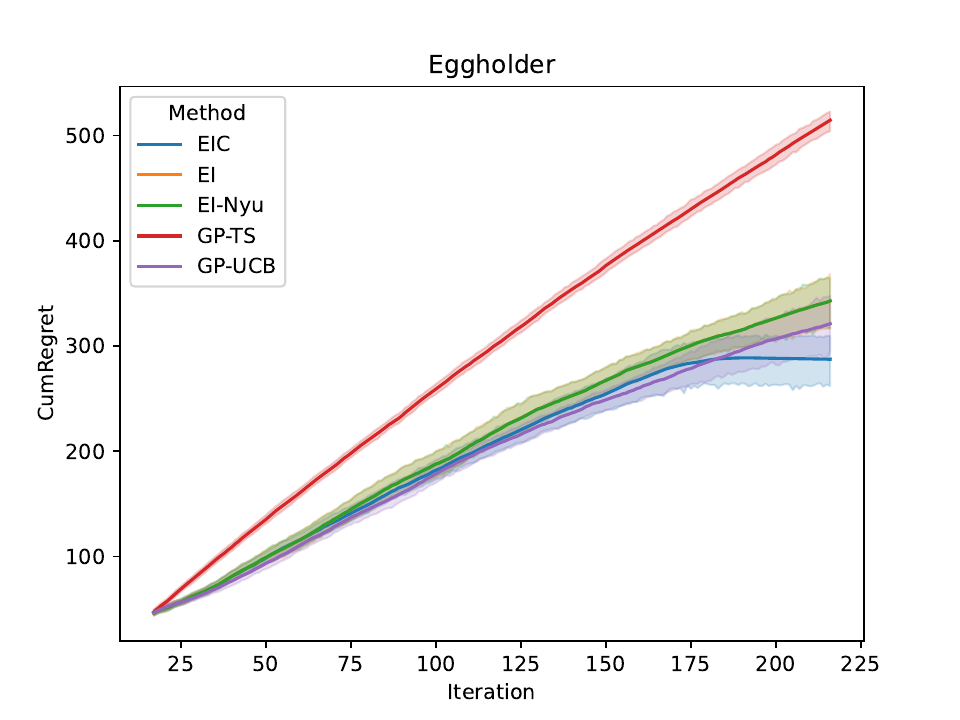}
		\caption{Eggholder-2}
		\label{fig:2b}
	\end{subfigure}
	
	\begin{subfigure}{0.5\linewidth}
		\includegraphics[width=\linewidth]{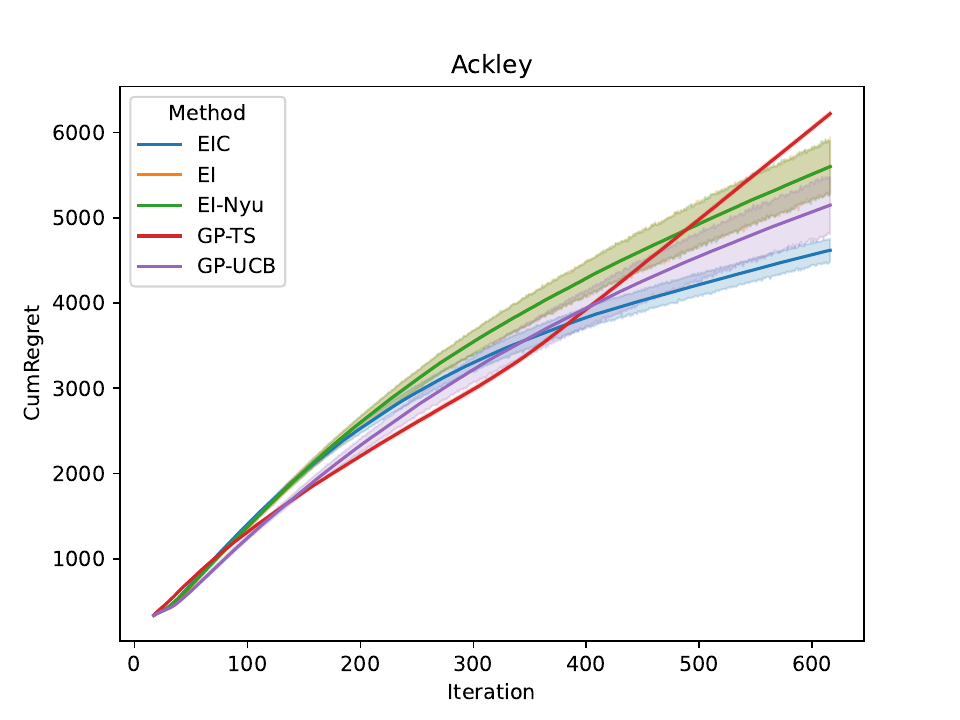}
		\caption{Ackley-2}
		\label{fig:2c}
	\end{subfigure}\hfill
	\begin{subfigure}{0.5\linewidth}
		\includegraphics[width=\linewidth]{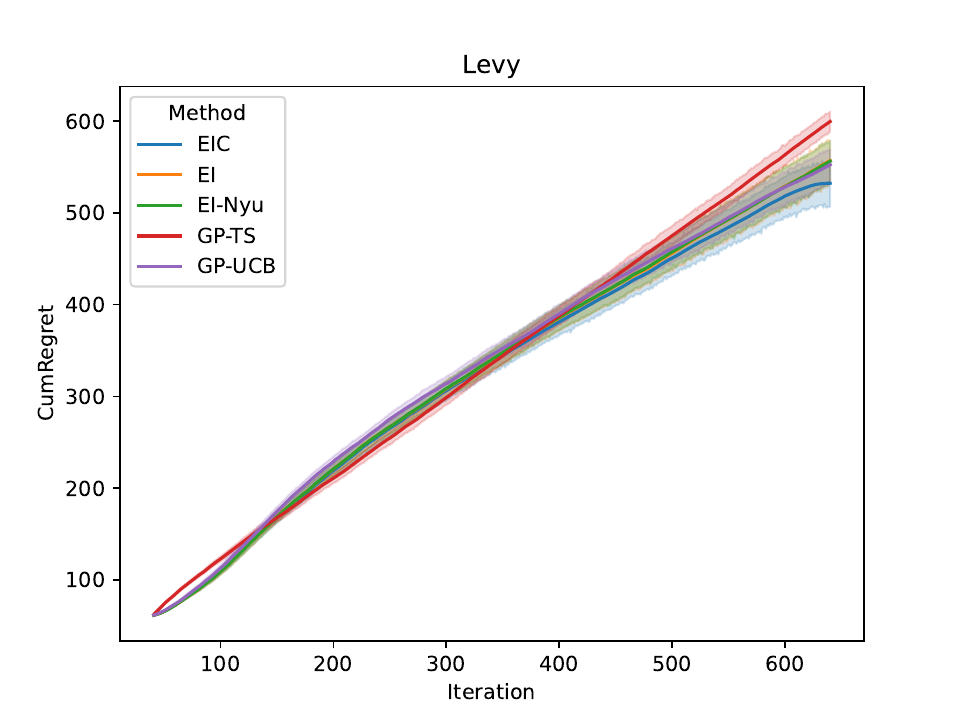}
		\caption{Levy-4}
		\label{fig:2d}
	\end{subfigure}
	
	\begin{subfigure}{0.5\linewidth}
		\includegraphics[width=\linewidth]{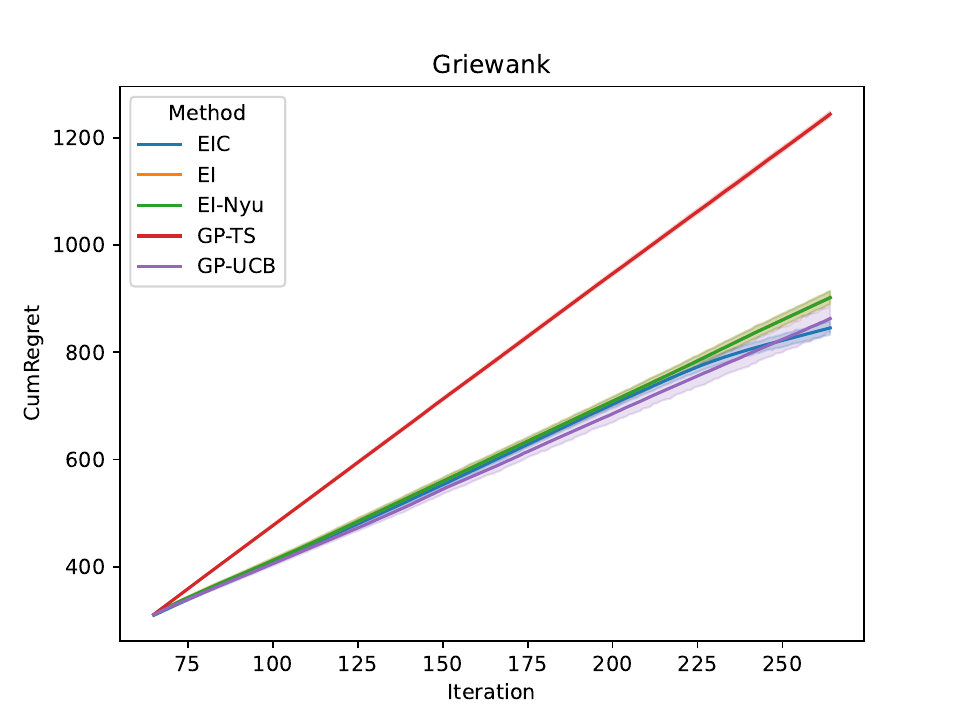} 
		\caption{Griewank-6}
		\label{fig:2e}
	\end{subfigure}\hfill
	\begin{subfigure}{0.5\linewidth}
		\includegraphics[width=\linewidth]{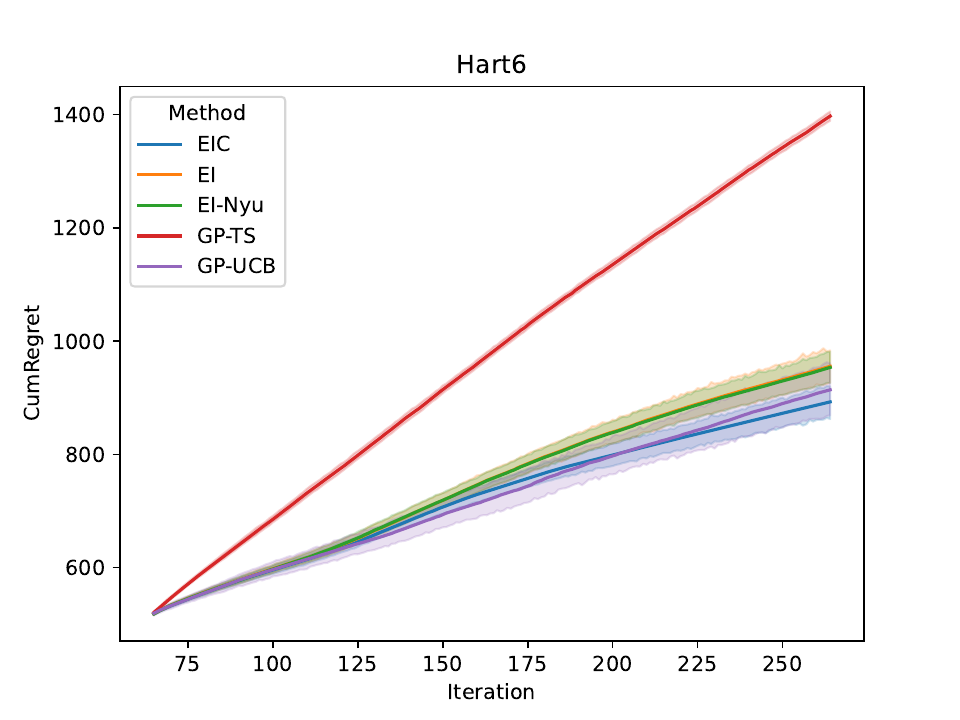}
		\caption{Hartmann-6}
		\label{fig:2f}
	\end{subfigure}
	\caption{Cumulative regret of different BO algorithms on six test functions.}
	\label{fig:2}
\end{figure}

In the experiments, we examine the numerical performance of our EIC algorithm using synthetic test functions (Section \ref{section:sync:experiments}) as well as a real-world dataset (Section \ref{section:real:experiments}).
For each test function/dataset, we generate $R = 100$ independent experiment trials, 
and the results are summarized in Figures \ref{fig:2} and \ref{fig:MLP}.
The solid line represents the average cumulative regret over these independent trials, 
with the shaded area showing  the corresponding 95\% confidence region.  
We use the GP prior with the mean function of $\mu \equiv 0 $ and the SE covariance kernel in all experiments.
The parameters  are estimated using the maximum likelihood method by \cite{SWN18}.
The total budget $N$ for each test function/dataset is set to $N=200 + n_0$,
where the number of initial design points is set at  $n_0 = 16, 36, 64$ for the $2, 4 ,6$--dimensional objective functions respectively.

We compare our proposed EIC algorithm with the acquisition functions of the traditional EI, GP-UCB \citep{Srinivas10}, and GP-TS \citep{CG17}.
We also compare with the algorithm proposed by \cite{Nguyen17} which we refer to as \emph{EI-Nyugen}.
The EI-Nyugen algorithm requires a user-specified threshold parameter $\kappa$, and selects the point that maximizes the acquisition function $\alpha_n(\bx)$ if the largest acquisition function value is not smaller than $\kappa$.
Otherwise, the point with the largest observed sample mean is selected.
We follow their suggestion and choose $\kappa = 10^{-4}$ in our experiments. 
%The EI-Nyugen is proposed by \cite{Nguyen17}, in which a pre-specified threshold $\kappa$ has been imposed. The algorithm selects a point that maximizes the acquisition function $\alpha_n(\bx)$ only if $\alpha_n(\bx) \geq \kappa$, Otherwise a point with the best sample mean is selected.

\begin{table}[!b]
	\centering
	\resizebox{\textwidth}{!}{  
		\begin{tabular}{llc}
			\hline
			\hline
			Functions &  $d$ & Equation  \\
			\hline
			Schwefel-2 &2 &  
			$\begin{gathered} 
				f(\mathbf{x})= -\frac{1}{274.3}\left( 418.9829 *2 -\sum_{i=1}^{2} w_{i} \sin \left(\sqrt{\left|w_{i}\right|}\right) - 838.57\right)\\  
				w_{i}=500x_i, i = 1, 2\\
				x_i \in [-1, 1],  i= 1, 2\\
				\mathbf{x}* = (0.8419,0.8419), f^* = 3.057
			\end{gathered}$\\
			\hline
			Eggholder-2 &2&
			$\begin{gathered} 
				f(\mathbf{x})= -\frac{1}{347.31}\left( -\left(w_{2}+47\right) \sin \left(\sqrt{\left|w_{2}+\frac{w_{1}}{2}+47\right|}\right)-w_{1} \sin \left(\sqrt{\left|w_{1}-\left(w_{2}+47\right)\right|}\right)- 1.96\right)\\
				w_{i}=512x_i, i = 1, 2\\
				x_i \in [-1.17, 1.17], i = 1, 2\\
				\mathbf{x}* = (1,0.7895), f^* = 2.769
			\end{gathered}$\\
			\hline
			Ackley-2&2& 
			$\begin{gathered} 
				f(\mathbf{x})=20 \exp \left(-0.2 \sqrt{\frac{1}{2} \sum_{i=1}^{2} x_{i}^{2}}\right)-\exp \left(\frac{1}{2} \sum_{i=1}^{2} \cos \left(2\pi x_{i}\right)\right)- 20- \exp (1)\\
				x_i \in [-32.768, 32.768], i= 1, 2\\
				\mathbf{x}* = (0,0), f^* = 0
			\end{gathered}$\\
			\hline
			Levy-4&4& 
			$\begin{gathered}
				f(\mathbf{x})=-\frac{1}{27.9}\left( \sin ^{2}\left(\pi w_{1}\right)+\sum_{i=1}^{3}\left(w_{i}-1\right)^{2}\left[1+10 \sin ^{2}\left(\pi w_{i}+1\right)\right]+\left(w_{4}-1\right)^{2}\left[1+\sin ^{2}\left(2 \pi w_{4}\right)\right] -42.55\right), \\
				w_{i}=1+\frac{x_{i}-1}{4}, \text { for all } i=1, \cdots, d \\
				x_i \in [-10, 10], i = 1, \cdots,4\\
				\mathbf{x}* = (1,1,1,1), f^* = 1.525
			\end{gathered}$ \\
			\hline
			Griewank-6&6&
			$\begin{gathered}
				f(\mathbf{x})= -\frac{1}{0.47}\left(\sum_{i=1}^{6} \frac{x_{i}^{2}}{4000}-\prod_{i=1}^{6} \cos \left(\frac{x_{i}}{\sqrt{i}}\right)+1 - 2.25\right)\\
				x_i \in [-50, 50], i = 1, \cdots,6\\
				\mathbf{x}* = (0,0,0,0,0,0), f^* = 4.787
			\end{gathered}$\\
			\hline
			Hartmann-6&6& 
			$\begin{gathered}
				f(\mathbf{x})=-\frac{1}{0.38}\left(-\sum_{i=1}^{4} \alpha_{i} \exp \left(-\sum_{j=1}^{6} A_{i j}\left(x_{j}-P_{i j}\right)^{2}\right) + 0.26\right)\\
				\alpha=(1.0,1.2,3.0,3.2)^{T}\\
				\mathbf{A}=\left(\begin{array}{cccccc}
					10 & 3 & 17 & 3.50 & 1.7 & 8 \\
					0.05 & 10 & 17 & 0.1 & 8 & 14 \\
					3 & 3.5 & 1.7 & 10 & 17 & 8 \\
					17 & 8 & 0.05 & 10 & 0.1 & 14
				\end{array}\right)\\
				\mathbf{P}=10^{-4}\left(\begin{array}{cccccc}
					1312 & 1696 & 5569 & 124 & 8283 & 5886 \\
					2329 & 4135 & 8307 & 3736 & 1004 & 9991 \\
					2348 & 1451 & 3522 & 2883 & 3047 & 6650 \\
					4047 & 8828 & 8732 & 5743 & 1091 & 381
				\end{array}\right)\\
				x_i \in [0, 1], i = 1, \cdots, 6\\
				\mathbf{x}* = (0.20169,0.150011,0.476874,0.275332,0.311652,0.6573), f^* = 8.059
			\end{gathered}$ \\
			\hline
			\hline
		\end{tabular}%
	}
	\caption{\label{tab:funcs} List of test functions}
\end{table}%

\subsection{Synthetic Experiments}
\label{section:sync:experiments}

In the synthetic experiments, we consider six commonly used test functions: Schwefel-2, Eggholder-2, Ackley-2, Levy-4, Griewank-6, and Hartmann-6. 
Table \ref{tab:funcs} lists the mathematical expressions of these functions. 
Before implementing any BO algorithm, all test functions are standardised so that the function values have mean zero and standard deviation one. 
The standard deviation of the homogenous noise $\sigma$ is set as $0.1$. 

{ Figure 2 summarizes the results. 
	We first observe that the traditional EI and EI-Nyugen have nearly identical performance for all test functions. 
	This is probably due to the pre-specified threshold $\kappa$ in EI-Nyugen being too small,
	which leads to the acquisition function seldom, if ever, falling below $\kappa$ throughout the entire experiment. 
	Therefore, EI-Nyugen behaves identically to the traditional EI. The GP-TS does not peforms well across different functions propbably due to its high randomness.
	Our proposed EIC algorithm has the smallest mean cumulative regret for the Eggholder-2,  Griewank-6 and Harmann-6 test functions after $200 + n_0$ iterations. In addition, for these functions, the confidence regions of CumRegret for EIC overlap with those of UCB,  showing EIC is competitive with UCB. The confidence regions of CumRegret for EIC does not overlapped with EI, EI-Nyugen and GP-TS, hence the reduction on cumulative regret is statistically significant.
	For the Schwefel-2, GP-UCB has the smallest cumulative regret.  However the confidence regions of GP-UCB, EIC, EI-Nyugen and traditional EI overlap, indicating there is no statistical significance among these algorithms under 95\% confidence level.
	For the Ackley-2 and  the Levy-4 function, we use a bigger budget of $N = 600 + n_0$, since none of the compared algorithms is able to converge after $200 + n_0$ iterations.
	For the Ackley-2, GP-UCB has the smallest cumulative regret at the beginning, but it is outperformed by EIC after 350 iterations. For Levy-4 function, EI-Nyugen,  EI, GP-UCB, and EIC have similar performance, but EIC shows a tendency to converge after 600 iterations while other algorithm's cumulative regret values continue to grow. 
	These results show that our EIC algorithm performs competitively and consistently across a variety of test functions.
	%GP-TS has the largest cumulative regret for all test functions except for Levy-4. 
	%This may be attributed to the large randomness of the GP-TS algorithm. Specifically, in each iteration of GP-TS, the next evaluation point is decided based on one randomly generated sample from the GP posterior, and any bad evaluation point will significantly increase the cumulative regret.
}

\subsection{Real-world Experiment}
\label{section:real:experiments}

\begin{figure}[!t]
	\centering
	\includegraphics[width=0.7\linewidth]{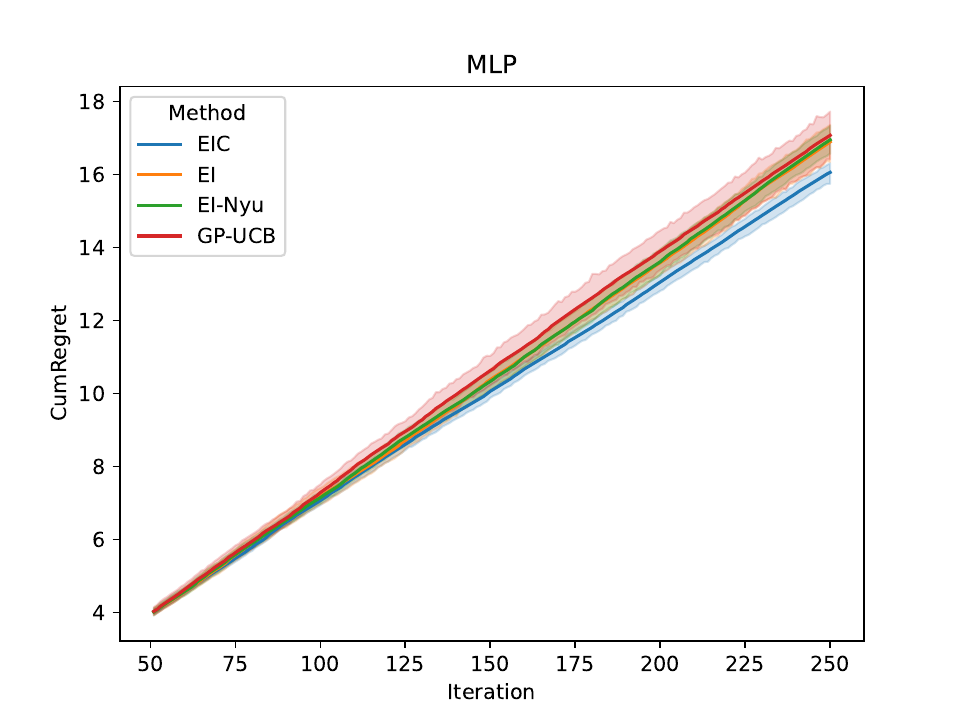}
	\caption{
		%	Cumulative regret performance of various BO algorithms on the neural network hyperparameter tuning problem.
		Cumulative regret of different BO algorithms in the neural network hyper-parameter tuning experiment.
	}
	\label{fig:MLP}
\end{figure}

\sloppy
%In this subsection, we test the optimization algorithms on a neural network hyperparameter tunning problem for a health care problem where we aim to build a classifier to identify whether a cancer is malignant or benign.  The example is used to simulate the clinical trial example in our introduction section.
In this section, we test the performance of various BO algorithms on a hyper-parameter tuning experiment for a neural network model, which is used in a healthcare application to classify whether a cancer is malignant or benign. 
%This experiment is used to simulate the motivating application of recommendation system mentioned in Section \ref{section:introduction}, for which cumulative regret is a more appropriate metric.
%For such application, it is essential that we could build an accurate classifier faster. 
We adopt the breast cancer Wisconsin dataset  (https://archive.ics.uci.edu/ml/datasets/breast+cancer+wisconsin+(diagnostic)) which contains 569 patients. 
Among them, 212 patients have malignant cancer and the other 357 have benign cancer. 
The dataset is further split into training and testing datasets with a ratio of 7:3. 
A one-hidden-layer multilayer perceptron (MLP) neural network model is trained to classify whether the patients have malignant or benign cancer based on 30-dimensional covariates. 
We consider the tuning of four hyper-parameters: the number of hidden units, batch size, learning rate, and learning rate decay coefficients. 
%The MLP is trained on the training dataset and then we fine-tune the four hyperparameters with Bayesian optimization algorithms by minimizing the classification error on the testing dataset. 
To evaluate a point (i.e., a $4$-dimensional vector specifying the values of the four hyper-parameters) selected by BO algorithms, we first train the MLP on the training dataset using the corresponding selected values of the hyper-parameters, and then evaluate the trained MLP on the testing dataset and report the classification accuracy as the corresponding observation.

{ 	In summary, this experiment involves a 4-dimensional optimization problem with the objective function $f$ representing the classification accuracy. 
Since the true maximizer $\mathbf{x}^*$ and the maximum value $f(\mathbf{x}^*)$ are unknown, we set the maximum value $f(\mathbf{x}^*)$ to be 1 (i.e., an accuracy of 100\%) in the calculation of the cumulative regret. 
Figure \ref{fig:MLP} summarizes the results for this experiment. 
We observe that 
for GP-UCB, EI-Nyugen and traditional EI, their confidence regions overlap and hence have very similar performance.
The EIC achieves the smallest cumulative regret after 150 iterations and the difference is significant as the its confidence regions do not overlap with those of other algorihthms.
}

\section{Conclusion}
\label{section:conclusion}
In this article, we propose the EIC algorithm which aims to enhance the performance of traditional EI under the evaluation metric of cumulative regret.
This is achieved by introducing an evaluation cost function that is compared against the acquisition function to balance the potential improvement to the cost of evaluation at each point. The algorithm is then designed to sample a point only if its acquisition function value exceeds its evaluation cost. 
We show that EIC can achieve  {a finite-time regret upper bound of $O\big(\sqrt{N} \gamma_{N} (\log N)^{1/2}\big)$ with high probability}, and use both synthetic and real-world experiments to demonstrate that EIC indeed achieves smaller cumulative regret compared to traditional EI as well as other commonly used BO algorithms.

We suggest here two extensions of BO for future work. 
The first is to consider the optimization in more complex domains, such as graphs, discrete sequences and trees.
The second is to handle the extra domain specific knowledge. For example in some nuclear safety applications we may know the optimum function value in advance, and how to incorporate such extra information still remains an open problem.

%{\bf Remarks} :
%\begin{itemize}
%	\item  
%	The performance of GP-UCB depends heavily on beta.
%	In Figure 2 a bad beta is used, and its performance is even worse than traditional EI.
%	Maybe we report GP-UCB with two different betas? 
%	One beta that corresponds to average performance of GP-UCB, and the other beta corresponds to the best performance of GP-UCB.
%	Argue that GP-UCB is very sensitive to beta? In practice it could be time consuming to tune beta.
%	(Try to avoid the criticism from reviewers).
%	
%	\item How to motivate that the real world dataset is suitable under the cumulative regret evaluation metric?
%	
%	\item Should we tune the incumbent of EIC also?
%	
%\end{itemize}

% Acknowledgements should go at the end, before appendices and references
\acks{This work is supported in part by the Ministry of Education, Singapore (Grant R-266-000-149-114).}


\newpage

\appendix

\section*{Appendix A. Proof of Supporting Lemmas}

% Note: in this sample, the section number is hard-coded in. Following
% proper LaTeX conventions, it should properly be coded as a reference:

%In this appendix we prove the following theorem from
%Section~\ref{sec:textree-generalization}:

\bigskip
   {
{\sc Proof of Lemma} \ref{lmCG}.
See Theorem 2 of \cite{CG17}. 
$\wbox$
}

\bigskip
   {
{\sc Proof of Lemma} \ref{lm2}.
Let $h(z) := z\Phi(z) +\phi(z)$.
We can write
\begin{eqnarray*}
	\alpha_{n}^{EI}(\bx)&  = & \omega_n \sigma_{n}(\bx) h\big( \tfrac{z_n(\bx)}{\omega_n} \big)  \text{ and }\\
	L_{n}(\bx)&  = &\omega_n \sigma_{n}(\bx) h\big( - \tfrac{ z_n(\bx) }{\omega_n} \big)/(N-n).
\end{eqnarray*}
Since $h(z) = z + h(-z)$, $\alpha_{n}^{EI}(\bx) \geq L_{n}(\bx)$ is equivalent to
\begin{equation} \label{znx}
	z_n(\bx) \geq -(N-n-1) \omega_n h\big( \tfrac{z_n(\bx)}{\omega_n} \big).
\end{equation} 
It suffices to show that (\ref{znx}) does not holds for $ z_n(\bx) < -\omega_n \sqrt{2 \log (N-n)}$, that is to show
\begin{equation} \label{lm3suffcondi}
	\tfrac{z_n(\bx)}{ \omega_n}  < -(N-n-1) h\big( \tfrac{z_n(\bx)}{\omega_n} \big) \text{ when } z_n(\bx) < -\omega_n \sqrt{2 \log (N-n)}.
\end{equation}
Check that (\ref{lm3suffcondi}) holds for $n = N - 1$.
Since $\Phi(x) > (-1/x+1/x^{3})\phi(x)$ for $x < 0$, we have 
\begin{equation*}
	h\big(\tfrac{z_n(\bx)}{\omega_n}  \big)  = \tfrac{z_n(\bx)}{\omega_n}   \Phi \big(\tfrac{z_n(\bx)}{\omega_n} \big) + \phi \big(\tfrac{z_n(\bx)}{\omega_n}  \big) 
	< \phi \big(\tfrac{z_n(\bx)}{\omega_n}  \big)/ (\tfrac{z_n(\bx)}{\omega_n}  )^2.
\end{equation*}
Hence to show (\ref{lm3suffcondi}), it suffices to show that for $1 \leq n \leq N - 2$,
\begin{equation} \label{phiz33}
	(N-n-1) \phi \big(\tfrac{z_n(\bx)}{\omega_n}  \big) < -\big(\tfrac{z_n(\bx)}{ \omega_n}\big)^3 \text{ when } z_n(\bx) < -\omega_n \sqrt{2 \log (N-n)}.
\end{equation}
Taking logarithm on both side of (\ref{phiz33}) give us
\begin{equation*} 
	\log \big( \tfrac{N-n-1}{\sqrt{2\pi}} \big) - \tfrac{1}{2} \big(\tfrac{z_n(\bx)}{ \omega_n}\big)^2 < 3 \log\big(-\tfrac{z_n(\bx)}{ \omega_n}\big).
\end{equation*}
Let $g(y) = \log \big( \tfrac{N-n-1}{\sqrt{2\pi}} \big) -\tfrac{1}{2} y^2 - 3 \log(-y)$. Check that $g^{\prime}(y) > 0$ for $y<0$ and 
\begin{equation*}
	g\big(-\sqrt{2 \log (N-n)}\big) =  \log \big( \tfrac{N-n-1}{N-n} \big) - \log(\sqrt{2\pi}) - \tfrac{3}{2}\log(2\log(N-n)) <0.
\end{equation*}
Hence (\ref{phiz33}) holds for for $1 \leq n \leq N - 2$.
$\wbox$
}

\bigskip
   {
{\sc Proof of Lemma} \ref{lm3}.
Define $h(z) := z\Phi(z) +\phi(z)$.
Let $z_{n} =  \tfrac{\mu_{n}(\bx)-\xi_n}{\sigma_{n}(\bx)}$ and  $q_{n} =  \tfrac{f(\bx)-\xi_n}{\sigma_{n}(\bx)}$. 
By Lemma \ref{lmCG}, we have 
$ |z_{n} - q_{n} | = | \tfrac{\mu_{n}(\bx) - f(\bx)}{\sigma_{n}(\bx)} | \leq \beta_{n}$
with probability at least $1-\delta$. \\
To show the upper bound, we have
\begin{eqnarray*}
	\alpha_{n}^{EI}(\bx) & =  & \omega_{n}\sigma_{n}(\bx) h(\tfrac{z_n}{\omega_{n}}) \\
	& \leq & \omega_{n}\sigma_{n}(\bx) h(\tfrac{q_n + \beta_{n}}{\omega_{n}}) \\
	& \leq & \omega_{n}\sigma_{n}(\bx) h(\tfrac{\max \lbrace 0, q_n \rbrace  + \beta_{n}}{\omega_{n}}) \\
	& \leq & \omega_{n}\sigma_{n}(\bx) [\tfrac{\max \lbrace 0, q_n \rbrace}{\omega_{n}}+\tfrac{\beta_{n}}{\omega_{n}} + 1] \\
	& \leq & I_n(\bx) + (\beta_{n} + \omega_{n}) \sigma_{n}(\bx),
\end{eqnarray*}
where the third inequality follows from $h(z) \leq z+ 1$ for $z\geq 0$.\\
To show the lower bound, note that
\begin{eqnarray} \label{EIlb1}
	\alpha_{n}^{EI}(\bx) & =  & \omega_{n}\sigma_{n}(\bx) h(\tfrac{z_n}{\omega_{n}}) \\ \nonumber
	& \geq  & z_n\sigma_{n}(\bx) \\ \nonumber
	& \geq & (q_n - \beta_{n})\sigma_{n}(\bx) \\ \nonumber
	& = & f(\bx) -\xi_n -\beta_{n}\sigma_{n}(\bx) \\ \nonumber
	& \geq &  I_n(\bx) - \beta_{n}\sigma_{n}(\bx),
\end{eqnarray}	
where the first inequality follows from $h(z) \geq z$ for all $z$.
Also, suppose that $f(\bx) -\xi_n \geq 0$, we have
\begin{eqnarray} \label{EIlb21}
	\alpha_{n}^{EI}(\bx) & \geq  & \omega_{n}\sigma_{n}(\bx) h(\tfrac{q_n -\beta_{n}}{\omega_{n}}) \\ \nonumber & \geq &  \omega_{n}\sigma_{n}(\bx) h(-\tfrac{ \beta_{n}}{\omega_{n}}) \\ 
\end{eqnarray} 
Combining (\ref{EIlb1}) and (\ref{EIlb21}) gives us
\begin{eqnarray} \label{EIlb22} 
	\alpha_{n}^{EI}(\bx) & \geq  & \frac{\omega_{n} h(-\tfrac{ \beta_{n}}{\omega_{n}})}{\omega_{n} h(-\tfrac{ \beta_{n}}{\omega_{n}}) + \beta_{n}}I_n(\bx) \\ \nonumber
	& =  & \frac{ h(-\tfrac{ \beta_{n}}{\omega_{n}})}{ h(\tfrac{ \beta_{n}}{\omega_{n}}) }I_n(\bx),
\end{eqnarray}  
where the last line follows from the fact that $h(z) = z + h(-z)$.
If $f(\bx) -\xi_n < 0$, (\ref{EIlb22}) still holds as  $\alpha_{n}^{EI}(\bx) > 0$. 
Hence Lemma \ref{lm3} follows.
$\wbox$
}

\bigskip
   {
{\sc Proof of Lemma} \ref{lm4}.
Suppose there exists some $1 \leq n_1 < n_2  < \cdots <n_K < N$ and $K \geq 1$ such that $f(\bx_{n_k +1}) - \xi_{n_k} \geq 0$ for $k = 1, \dots ,K$. 
Let $n_0 = 0$.
With probability at least $1-\delta$, we have 
\begin{eqnarray} \label{sumIn}
	\sum_{n=1}^{N-1} I_n(\bx_{n+1}) & = &\sum_{n=1}^{N-1}  \big(f(\bx_{n+1}) - \xi_n \big) \mathbf{1}_{\lbrace f(\bx_{n+1}) \geq \xi_n \rbrace} \\ \nonumber
	& =  &  \sum_{k = 1}^{K} \big( f(\bx_{n_k+1}) - \xi_{n_k} \big) \\ \nonumber
	& \leq & \sum_{k = 1}^{K} \big( f(\bx_{n_k+1}) - \mu_{n_k}(\bx_{n_{k-1}+1}) \big) \\ \nonumber
	& \leq & \sum_{k = 1}^{K} \big( f(\bx_{n_k+1}) - f(\bx_{n_{k-1}+1}) + \beta_{n_k} \sigma_{n_k}(\bx_{n_{k-1}+1}) \big) \\ \nonumber
	& = &   f(\bx_{n_K+1}) - f(\bx_{n_0+1})+ \sum_{k = 1}^{K}\beta_{n_k} \sigma_{n_k}(\bx_{n_{k-1}+1})
\end{eqnarray}
The first inequality of (\ref{sumIn}) follows from the fact that $\xi_{n_k} = \max_{1 \leq i \leq n_k}\mu_{n_k}(\bx_{i}) \geq  \mu_{n_k}(\bx_{n_{k-1}+1})$ and the second inequality follows from Lemma 1. 
By Lemma \ref{lemsubsigma}, the posterior variance is monotonically decreasing in $n$, and we have
\begin{eqnarray} \label{lm4arg3}
	\sum_{k = 1}^{K}\beta_{n_k} \sigma_{n_k}(\bx_{n_{k-1}+1})  & \leq & \beta_{N} \sum_{k = 1}^{K} \sigma_{n_k}(\bx_{n_{k-1}+1}) \\ \nonumber
	& \leq & \beta_{N} \sum_{k = 1}^{K} \sigma_{n_{k-1}}(\bx_{n_{k-1}+1}) \\  \nonumber
	& \leq & \beta_{N} \sum_{n = 0}^{N-1} \sigma_{n}(\bx_{n+1}) 
\end{eqnarray}
Hence Lemma \ref{lm4} follows from (\ref{sumIn}),  (\ref{lm4arg3}) and Lemma \ref{lem2B}.
$\wbox$
}

\bigskip
   {
{\sc Proof of Lemma} \ref{lm5}.
Let $y_n = f(\bx_n)+ \epsilon_n$ with $\epsilon_n 
\sim N(0, \lambda^2 \omega_{n-1}^2)$.
Let $F_N = \big(f(\bx_1),f(\bx_2)\dots,f(\bx_N) \big)^T$ and $Y_N = (y_1, y_2, \dots, y_N)^T$.
The mutual information between $Y_N$ and $F_N$ can be written as
\begin{equation}
	I(Y_N;F_N) = H(Y_N) - H(Y_N|F_N),
\end{equation} 
where $H(X)$ denote the differential entropy of $X$ and $H(X|Y)$ denote the conditional (differential) entropy of $X$ given $Y$.
By the chain rule for entropy, 
\begin{equation} \label{HYn}
	H(Y_N) = \sum_{n = 1}^{N} H(y_n|y_{n-1},\dots,y_1) = \frac{1}{2} \sum_{n = 1}^{N} \log\big(2\pi e \omega_{n-1}^2 (\sigma_{n-1}^2(\bx_{n})+\lambda^2) \big),
\end{equation}
where the last equality follows since $y_n|y_{n-1},\dots,y_1 \sim N \big( \mu_{n-1}(\bx_{n}), \omega_{n-1}^2 (\sigma_{n-1}^2(\bx_{n})+\lambda^2)\big)$.
Moreover, conditioning on $F_N$, $Y_N$ follows a multivariate normal distribution with covariance matrix $\Sigma = \lambda^2  \text{diag}(\omega_0^2, \omega_1^2,\dots, \omega_{N-1}^2 )$. Hence
\begin{equation} \label{HYnFn}
	H(Y_N|F_N) =  \frac{1}{2}\log \big( (2\pi e)^N \text{det}(\Sigma ) \big) = \frac{1}{2} \sum_{n = 1}^{N} \log ( 2\pi e \lambda^2 \omega_{n-1}^2).
\end{equation}
Combining (\ref{HYn}) and (\ref{HYnFn}) gives us
\begin{equation*}
	I(Y_N;F_N) = \frac{1}{2} \sum_{n = 1}^{N} \log \big(1 + \tfrac{\sigma_{n-1}^2(\bx_{n})}{\lambda^2} \big) =  \frac{1}{2} \sum_{n = 0}^{N-1} \log \big(1 + \tfrac{\sigma_{n}^2(\bx_{n+1})}{\lambda^2} \big).
\end{equation*} 
By the definition of maximum information gain, $\gamma_{N} \geq I(Y_N;F_N)$. 
Since $\log(1+x) \geq x$ for all $x > 0 $, we have
\begin{equation*}
	\gamma_{N} \geq \frac{1}{2} \sum_{n = 0}^{N-1} \log \big(1 + \tfrac{\sigma_{n}^2(\bx_{n+1})}{\lambda^2} \big) \geq \frac{1}{2} \sum_{n = 0}^{N-1} \tfrac{\sigma_{n}^2(\bx_{n+1})}{\lambda^2}.
\end{equation*}
Rearranging the terms gives us Lemma \ref{lm5}.
$\wbox$
}

\bigskip
   {
{\sc Proof of Lemma} \ref{lem2B}.
By (\ref{RPP}), the Cauchy-Schwarz inequality and (\ref{rkhsb}),
\begin{eqnarray} \label{lm2Barg1}
	f(\bx) - f(\bx^\prime) & = &\langle f, k(\cdot,\bx) \rangle_{\mathcal{H}_{k}} -  \langle f, k(\cdot,\bx_{n_0}) \rangle_{\mathcal{H}_{k}} \\ \nonumber
	& = &\langle f, k(\cdot,\bx) -  k(\cdot,\bx^\prime) \rangle_{\mathcal{H}_{k}} \\ \nonumber
	& \leq  &\Vert f \Vert_{\mathcal{H}_k} \Vert k(\cdot,\bx) -  k(\cdot,\bx^\prime) \Vert_{\mathcal{H}_k} \\ \nonumber
	& \leq  & B \Vert k(\cdot,\bx) -  k(\cdot,\bx^\prime) \Vert_{\mathcal{H}_k}.
\end{eqnarray}
Since $k$ is isotropic, $k(\bx,\bx) = 1$ and $k(\bx,\bx^\prime) > 0$, we have
\begin{eqnarray} \label{lm2Barg2}
	& &\Vert k(\cdot,\bx) -  k(\cdot,\bx^\prime) \Vert_{\mathcal{H}_k}\\ \nonumber
	& =  &\sqrt{\langle k(\cdot,\bx) -  k(\cdot,\bx^\prime), k(\cdot,\bx) -  k(\cdot,\bx^\prime) \rangle_{\mathcal{H}_{k}} } \\ \nonumber
	& = & \sqrt{2 - k(\bx,\bx^\prime) } \leq \sqrt{2}.
\end{eqnarray}
Combining (\ref{lm2Barg1}) and (\ref{lm2Barg2}) give us the result.
$\wbox$
}

\bigskip
   {
{\sc Proof of Lemma} \ref{lemsubsigma}.
Recall that $X = (\bx_1,\dots,\bx_n)^T$, $k_{\bx X} = k_{X \bx }^T = \big( k(\bx_1,\bx),\dots,k(\bx_n,\bx) \big)$ and the $(i,j)$ entry of $K_{XX}$ is $k(\bx_{i},\bx_{j})$.
Define the leave-one-out vector $X^{(-n)} = (\bx_1,\dots,\bx_{n-1})^T$ and $k_{\bx X^{(-n)}} = k_{X^{(-n)} \bx }^T = \big( k(\bx_1,\bx),\dots,k(\bx_{n-1},\bx) \big)$. we can express
\begin{equation*}
   K_{XX}+\lambda^2 I_n = 
   \begin{pmatrix}
   \begin{matrix}
   	K_{X^{(-n)}X^{(-n)}}+ \lambda^2 I_{n-1}& k_{X^{(-n)}\bx_n} \\
   	k_{\bx_n X^{(-n)}} & k(\bx_n,\bx_n) +\lambda^2
   \end{matrix}
   \end{pmatrix}.
\end{equation*}
Using the block matrix inversion formula, we have
\begin{equation} \label{kmatrixinv}
	(K_{XX}+\lambda^2 I_n)^{-1} = 
	\begin{pmatrix}
		\begin{matrix}
			A^{-1} + \rho^{-1} A^{-1}BB^{T}A^{-1}& -\rho^{-1} A^{-1}B \\
			-\rho^{-1} B^{T}A^{-1} & \rho^{-1}
		\end{matrix}
	\end{pmatrix},
\end{equation}
where $	A^{-1} = (K_{X^{(-n)}X^{(-n)}}+ \lambda^2 I_{n-1})^{-1}$, $B = k_{X^{(-n)}\bx_n}$ and $\rho = (k(\bx_n,\bx_n) +\lambda^2) - B^TA^{-1}B = \sigma_{n-1}(\bx_{n}) + \lambda^2$ is the Schur complement. By (\ref{kmatrixinv}) and the block matrix multiplication formula:
\begin{eqnarray*}
	\sigma_{n}^2(\bx) & = & k(\bx,\bx) - k_{\bx X} (K_{XX}+\lambda^2 I_n)^{-1} k_{X \bx } \\ \nonumber
	& = & k(\bx,\bx) - \begin{pmatrix}
		\begin{matrix}
			k_{\bx X^{(-n)}} \;\; k(\bx,\bx_n) 
		\end{matrix}
	\end{pmatrix} 
	\begin{pmatrix}
		\begin{matrix}
			A^{-1} + \rho^{-1} A^{-1}BB^{T}A^{-1}& -\rho^{-1} A^{-1}B \\
			-\rho^{-1} B^{T}A^{-1} & \rho^{-1}
		\end{matrix}
	\end{pmatrix}
	\begin{pmatrix}
		\begin{matrix}
			k_{X^{(-n)}\bx} \\
			k(\bx_n,\bx) 
		\end{matrix}
	\end{pmatrix} \\ \nonumber
	& = & k(\bx,\bx) - k_{\bx X^{(-n)}} A^{-1} k_{X^{(-n)}\bx} - \rho^{-1} k_{\bx X^{(-n)}} A^{-1}BB^{T}A^{-1} k_{X^{(-n)}\bx}  \\ \nonumber
     & \quad & \quad  	+ \; \rho^{-1} k_{\bx X^{(-n)}}  A^{-1}B k(\bx_n,\bx) + \rho^{-1}   k(\bx,\bx_n) B^T  A^{-1} k_{X^{(-n)} \bx} - \rho k(\bx,\bx_n)k(\bx_n,\bx) \\ \nonumber
     & = & \sigma_{n-1}^2(\bx) - \rho^{-1}\big( k^2(\bx,\bx_n) -2  k(\bx,\bx_n) B^T  A^{-1} k_{X^{(-n)}\bx} +   (B^{T}A^{-1} k_{X^{(-n)}\bx})^2 \big) \\
     & = & \sigma_{n-1}^2(\bx)  - \rho^{-1}  \big( k(\bx,\bx_n) - B^{T}A^{-1} k_{X^{(-n)}\bx}\big)^2 \\ \nonumber
     & = & \sigma_{n-1}^2(\bx) - \frac{k_{n-1}(\bx,\bx_n)}{\sigma_{n-1}(\bx_{n}) + \lambda^2}.  \text{ $\wbox$ }
\end{eqnarray*}  
}

%\medskip
%   {
%	\begin{lem}		
%		Let $\xi_n := \mu_{n-1}^* = \max_{1 \leq i \leq n-1} \mu_{n-1}(\bx_{i})$ and let $\bx_m := \argmin_{1 \leq i \leq n_0} \Vert \bx^* - \bx_i \Vert_{\mathbf{h}}$.
%		With probability at least $1 - \delta$, we have
%		\begin{equation*}
%			f(\bx^*) - \xi_n \leq B\sqrt{2- 2\exp(-\tfrac{d}{8M^2})} + \beta_n \sigma_{n-1}(\bx_m).
%		\end{equation*}
%	\end{lem}
%	{\sc Proof}. 	After the initial experiment, the algorithm has sampled $\bx_1,\dots,\bx_{n_0}$ according to (\ref{idp}). 
%	We have
%	\begin{equation} \label{z2sd}
%		\Vert \bx^* - \bx_m \Vert_{\mathbf{h}} = \min_{1 \leq i \leq n_0}  \Vert \bx^* - \bx_i \Vert_{\mathbf{h}} \leq  \sqrt{(\tfrac{1}{2M})^2 + \cdots + (\tfrac{1}{2M} )^2} = \sqrt{\tfrac{d}{4M^2}}.
%	\end{equation}
%	By (\ref{RPP}), the Cauchy-Schwarz inequality and (\ref{rkhsb}) again,
%	\begin{eqnarray} \label{z2inigap}
%		f(\bx^*) - \xi_n & = & f(\bx^*) - \max_{1 \leq i \leq n-1} \mu_{n-1}(\bx_{i})  \\ \nonumber
%		&  \leq & f(\bx^*) - \mu_{n-1}(\bx_m) \\ \nonumber
%		&  \leq & f(\bx^*) - f(\bx_m) + \beta_n \sigma_{n-1}(\bx_m)\\ \nonumber
%		& \leq & B \sqrt{2 -2 k(\bx^*,\bx_m)} + \beta_n \sigma_{n-1}(\bx_m) \\ \nonumber
%		& = & B  \sqrt{2 - 2 \exp \big(-\tfrac{1}{2} \Vert \bx^* - \bx_m \Vert_{\mathbf{h}}^2 \big) } + \beta_n \sigma_{n-1}(\bx_m) \\ \nonumber
%		& \leq & B  \sqrt{2 - 2 \exp \big(-\tfrac{d}{8M^2} \big)} + \beta_n \sigma_{n-1}(\bx_m), 
%	\end{eqnarray}
%	where the last inequality follows from (\ref{z2sd}).
%	$\wbox$
%}

\end{document}